%% file: sample-manuscript.tex
\documentclass[manuscript,screen,table]{acmart}
\AtBeginDocument{%
  \providecommand\BibTeX{{%
    \normalfont B\kern-0.5em{\scshape i\kern-0.25em b}\kern-0.8em\TeX}}}

\setcopyright{none}
\copyrightyear{2023}

\acmYear{2023}
\acmDOI{XXXXXXX.XXXXXXX}

\setcitestyle{nocompress}
\usepackage{graphicx}
\usepackage{booktabs}
\usepackage{multirow}
\usepackage{tikz}
\usetikzlibrary{arrows.meta,decorations.pathmorphing,backgrounds,positioning,fit,petri}
\usepackage[edges]{forest}
\usepackage{subcaption}
\usepackage{calc}
\usepackage{tabularx}
\usepackage{vcell}
\usepackage{booktabs,siunitx}

\usepackage [english]{babel}
\usepackage [autostyle=true, english = american]{csquotes}
\MakeOuterQuote{"}

\definecolor{rowzebra}{HTML}{f5f7fa}
\begin{document}

\title{Racial Bias within Face Recognition: A Survey}

\author{Seyma Yucer}
\affiliation{%
  \institution{Durham University}
  \city{Durham}
  \country{UK}
}

\author{Furkan Tektas}
\affiliation{%
  \institution{BuboAI}
  \city{Middlesbrough}
  \country{UK}}

\author{Noura Al Moubayed}
\affiliation{%
  \institution{Durham University}
  \city{Durham}
  \country{UK}
}

\author{Toby P. Breckon}
\affiliation{%
 \institution{Durham University}
 \city{Durham}
 \country{UK}}

\renewcommand{\shortauthors}{Yucer and Tektas, et al.}


\input{tex/0-Abstract.tex}

\maketitle
\input{tex/1-Introduction.tex}
\input{tex/2-Preliminaries.tex}

\input{tex/3-TowardsRacialGroupFairness.tex}
\input{tex/4-RacialBiasWFaceRecognition.tex}

\input{tex/5-Conclusions.tex}

\input{tex/references.tex}
\appendix

\end{document}

%% file: tex/0-Abstract.tex
\begin{abstract}

Facial recognition is one of the most academically studied and industrially developed areas within computer vision where we readily find associated applications deployed globally. This widespread adoption has uncovered significant performance variation across subjects of different racial profiles leading to focused research attention on racial bias within face recognition spanning both current causation and future potential solutions. In support, this study provides an extensive taxonomic review of research on racial bias within face recognition exploring every aspect and stage of the face recognition processing pipeline. Firstly, we discuss the problem definition of racial bias, starting with race definition, grouping strategies, and the societal implications of using race or race-related groupings. Secondly, we divide the common face recognition processing pipeline into four stages: image acquisition, face localisation, face representation, face verification and identification, and review the relevant corresponding literature associated with each stage. The overall aim is to provide comprehensive coverage of the racial bias problem with respect to each and every stage of the face recognition processing pipeline whilst also highlighting the potential pitfalls and limitations of contemporary mitigation strategies that need to be considered within future research endeavours or commercial applications alike. 
\end{abstract}

%% file: tex/1-Introduction.tex
\vspace{-.4cm}
\section{Introduction}


Over several decades, the objective of developing face recognition systems has gathered significant pace across research, and industry alike \cite{du2020elements,ali2021classical,wang2021deep}. Companies, nonprofits, and governments have deployed an increasing number of face recognition systems to make autonomous decisions for millions of users \cite{kortli2020face} across various application areas, such as within employment decisions, public security, criminal justice, law enforcement surveillance, airport passenger screening, and credit reporting \cite{biometricrecognition,amos2016openface}. However, such wide-scale adoption within real-world scenarios heightens public concern about their potential for abuse and the adverse effect of face recognition may have on some individuals due to the presence of bias \cite{garcia2019harms,srinivas2019face}. The most prevalent problem pertaining to such bias arises within the race and race-related groupings and is referred to as racial bias within face recognition \cite{grother2019face}.

However, the presence of racial bias within face recognition is not a new thing and is not in itself limited to technological means. \textit{Own-race bias} has been previously established in psychology \cite{meissner2001thirty} by showing that humans are less capable of recognising faces from other races than their own. The prolonged societal experience humans generally have with their own-race, especially during their formative years with biological family members, results in biased human perceptual expertise. More specifically, \cite{hills2006short} showed how the use of face feature descriptors varies across participants from different racial groupings. For example, it shows that darker skin tone participants use face outline, eye size, eyebrows, chin and ears, while lighter skin tone participants use hair colour, texture, and eye colour. Overall, it concludes that lighter skin tone participants use less varied descriptors than darker skin tone participants \cite{hills2006short}. Similar to the \textit{own-race bias}, the conversely named \textit{other-race effect} is also studied by a series of studies in social psychology \cite{anzures2013developmental,rhodes2009race} to establish social implications of biased face processing and feature selection of humans in erroneous jury decisions, eyewitness identification.

Accordingly, the first technological study \cite{phillips2011other} to explore the \textit{other-race effect} within the context of face recognition algorithms was developed by East Asian and Western-based research groups that inherently use datasets gathered locally. The study demonstrates that algorithms trained on a locally gathered face datasets from the Western-based group achieve superior performance on Caucasian faces when compared to performance on East Asian faces, and vice versa. Further studies provide extensive evidence about the influence of demographics, including race, gender and age on both commercial and non-commercial face recognition algorithm performance \cite{klare2012face,o2012demographic}. Subsequently, the Gender Shades study \cite{buolamwini2018gender} drew significant attention to gender and skin tone bias within commercial algorithms for gender classification by revealing a 34\% performance discrepancy between darker skin tone female and lighter skin tone male subjects. Consequently, growing research has emerged to understand and mitigate racial bias within face recognition \cite{karkkainen2021fairface,menezes2021bias,yucer2020exploring}. These efforts and associated evidence of bias have forced several commercial and academical research to withdraw products, algorithms, or datasets due to the differing forms of disparities, distortions or biases \cite{shiaeles2021facebook,menn2019microsoft,castelvecchi2020facial}.

However, face recognition remains a long-standing research topic and a common use case within computer vision that comprises multiple stages of processing, a multitude of downstream tasks and large-scale face recognition datasets in order to achieve high accuracy. With the availability of such large-scale data resources and the advent of Deep Convolutional Neural Networks (DCNN), the accuracy of face recognition algorithms has now excelled the perceived accuracy requirements for use by the general populous. However, every stage of face recognition, from initial face image acquisition to final performance evaluation, requires attention and investigation to address racial bias, which may otherwise result in disparate outcomes across a diverse user population. Unfortunately, despite the increasing attention to racial bias within face recognition, we are yet to see truly collaborative or tractable solutions emerge from the global research base that could readily address these issues in real-world system deployments \cite{wang2019racial,sixta2020fairface,gebru2021datasheets,yucer2022measuring}. Moreover, face data itself is a private biometric capable of identifying a given individual based on their appearance alone, giving rise to obvious operational privacy and ethical concerns in relation to its processing \cite{biometricreveal}. 
Although previous surveys on algorithmic bias and fairness in machine learning \cite{drozdowski2020demographic,mehrabi2021survey,orphanou2022mitigating} and face recognition in computer vision and biometrics \cite{kortli2020face,wang2021deep} exist, many aspects remain under-studied in relation to the specifics of racial bias within face recognition. 

On the other hand, face recognition is a fast emerging field of research and applications alike that spans multiple more traditional fields, including machine learning, biometrics, statistics, sociology, and psychology. Therefore, we commonly find that aspects of the problem definition, in addition to the race conceptualisation and race-related performance evaluation methodologies, need to be clarified and ideally standardised. Which stages, operations and decisions in face recognition are prone to bias, and how incorrect solutions to addressing the bias issue can cause additional areas of concern that need to be highlighted in order to maximise the effectiveness of future research in this area. In this survey, we take face recognition as the central concept of our review and aim to provide coverage of all the aspects of racial bias within each stage of the face recognition processing pipeline, with additional supporting material spanning fundamental concepts from related fields.

\begin{figure}[t]
\centering
\resizebox{\linewidth}{!}{
\input{figures/taxonomy.tikz}
}
\vspace{-0.6cm}
\caption{Taxonomy of sections in Racial Bias within Face Recognition Survey.}\label{fig:taxonomy}
\vspace{-0.5cm}
\end{figure}
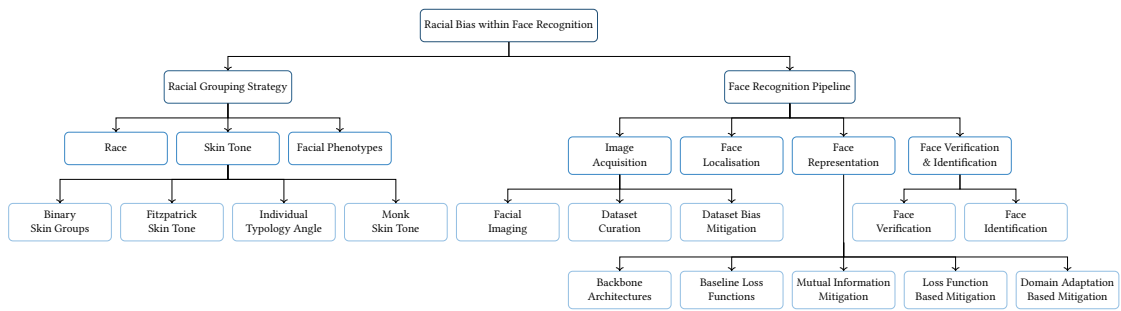

The primary purpose of this study is to both summarise the current state of the art and to give a comprehensive critical review of prior research on the topic of racial bias within face recognition. In addition, we aim to make the reader pertinently aware as to the subtleties, and potential areas of ambiguity, with regard to how the racial bias problem within face recognition itself is defined. Furthermore we aim to identify which parts of the problem have been studied effectively to date and which directions remain open for future contributions to mitigate racial bias within the face recognition domain. In particular, the survey aims to systematically review each of the stages that are commonplace within contemporary face recognition processing pipelines from a perspective of the potential for racial bias impact: image acquisition (for both dataset collation and deployment), face localisation, face representation, face verification and identification (final decision-making) (see Figure \ref{fig:taxonomy}, right).

On this basis, we present this survey based on our taxonomy of prior work in the field and its contribution to the current state of the art (Figure \ref{fig:taxonomy}).
Subsequently, we formalise the problem definition with the corresponding evaluation and fairness criteria (Section \ref{sec:2}).
Next, we discuss standard race and race-related grouping terminology under three categories; race, skin tone and facial phenotypes (Section \ref{sec:3}). This discussion provides an information spectrum from grouping definitions to their adoption to the associated processing of racial groupings used in literature studies. Consequently, we provide a general development schema for face recognition systems and summarise the prior work in the field by aligning it to each development stage (Section \ref{sec:4}). Within this section (Section \ref{sec:4}), we firstly give an outline description of the general face recognition processing pipeline using consistent notions and symbols. Secondly, we cover image and dataset acquisition processes for face recognition showing the risks and investigations within this stage. Thirdly, we extend our analysis to face localisation as it is a mandatory stage where the possible biased localisation results propagate within the following face recognition stages. Penultimately, in the face representation stage, we categorise the proposed racial bias mitigation approaches based on machine learning techniques. Finally, we cover face identification and verification tasks and show the impact of the methodological decisions effects on racial bias.
Consequently, we summarise the main critical points of the work and highlight the essential steps that need to be considered within any future research endeavours or commercial applications that aim to mitigate bias or develop fairer face recognition systems (Section \ref{sec:5}).

%% file: figures/taxonomy.tikz
\definecolor{l1}{HTML}{225076}
\definecolor{l2}{HTML}{285E89}
\definecolor{l3}{HTML}{3987c5}
\definecolor{l4}{HTML}{92BEE2}
\forestset{
  myrect/.style={
    draw=#1,
    rectangle,
    rounded corners=3pt,
    minimum height=2.5em,
    minimum width=2.5cm,
    font=\small,
    align=center,
  },
  myrect/.default=gray
  }

\begin{forest}
    for tree={
        myrect,
        align=center,
        edge={ ->},
        edge path={
            \noexpand\path[\forestoption{edge}]
            (!u.parent anchor) -- +(0,-10pt) -|   
            (.child anchor)\forestoption{edge label};
        },
        parent anchor=south,
        child anchor=north,
        l sep+=1em,
        s sep+=0em,
        delay={
            where content={}{
              shape=coordinate,
            }{},
      },
    },
    before typesetting nodes={
        where content={}{coordinate}{},
    }
    [Racial Bias within Face Recognition,myrect=l1 
        [Racial Grouping Strategy,myrect=l2,tier=l2
            [Race,myrect=l3,tier=l3 ]
            [Skin Tone,myrect=l3,tier=l3              
                [Binary\\Skin Groups,myrect=l4,tier=l4]
                [Fitzpatrick\\Skin Tone,myrect=l4,tier=l4]
                [Individual\\Typology Angle,myrect=l4,tier=l4]
                [Monk\\Skin Tone,myrect=l4,tier=l4]
            ]
            [Facial Phenotypes,myrect=l3,tier=l3 ]
        ]
        [Face Recognition Pipeline,myrect=l2,tier=l2           
                [Image\\Acquisition,myrect=l3,tier=l3 
                    [Facial\\Imaging,myrect=l4,tier=l4]
                    [Dataset\\Curation,myrect=l4,tier=l4]
                    [Dataset Bias\\Mitigation,myrect=l4,tier=l4]
                ]
            [Face\\Localisation,myrect=l3,tier=l3 ]
            [Face\\Representation,myrect=l3,tier=l3  [,l*=2.5,coordinate,edge={-}
                        [Backbone\\Architectures,myrect=l4,tier=l5]
                        [Baseline Loss\\Functions,myrect=l4,tier=l5]
                        [Mutual Information\\Mitigation,myrect=l4,tier=l5]
                        [Loss Function\\Based Mitigation,myrect=l4,tier=l5]
                        [Domain Adaptation\\Based Mitigation,myrect=l4,tier=l5]
            ]]
            [Face Verification\\ \& Identification,myrect=l3,tier=l3                 
                [Face\\Verification,myrect=l4,tier=l4]
                [Face\\Identification,myrect=l4,tier=l4]
            ]
        ]
    ]
\end{forest}

%% file: tex/2-Preliminaries.tex
\vspace{-.4cm}
\section{Preliminaries}
\label{sec:2}

Statistical methods are essential for supervised learning problems, including face recognition, which concerns generating a representative and distinctive feature embedding vector $z$ for a subject $y$ given an observed face image $x$. A mapping function $f^*$ is a particular function among infinite function space $\Omega$ $(f^* \in \Omega )$ that provides optimal performance over a given training dataset $D_{train}$. Preferring certain functions over others is denoted as inductive bias in the seminal work by Mitchell \cite{mitchell1980need} and remains a central concept in statistical learning theory. The expression \textit{inductive bias} (also known as learning bias) refers to the optimal selection process of $f^*$. Due to its importance for generalisation on unseen large-scale datasets, inductive bias is essential for any genre of machine learning approach.
On the other hand, the broader societal, historical meaning of the term \textit{bias} instead refers to the unfair treatment of a subset of the populous based on their origins, ethnicity or ideology. While \textit{inductive bias} is necessary for model generalisation, \textit{societal bias} implies negative implications that should ideally be avoided \cite{hellstrom2020bias}. In order to avoid the obvious potential for confusion, the prior work of \cite{barocas2017fairness} prefers to use fairness instead of bias when referring to aspects of demographic criteria in both statistics and machine learning. Subsequently, research on algorithmic fairness and statistical bias has introduced various formal definitions of fairness, and their relationships with each other \cite{Dwork2012,barocas2017fairness,kusner2017counterfactual}. Before we fully detail these fairness criteria, we first provide a brief explanation of a generic face representation learning and evaluation pipeline to facilitate the introduction of the required notation, which we will subsequently use for the remainder of this review.

A face recognition system comprises a training set $D_{train}$ and a test set $D_{test}$ where any of the datasets can be defined as $D=\{X, Y\}$ where $X=\{x_1,x_2,..,x_N\}$ is a set of face images and $Y=\{y_1,y_2,..,y_N\}$ is a set of subject identity labels corresponding the face images where $N$ is the total number of images. The total number of unique subject identity labels is $n$ and is smaller than $N$.
In addition, in order to measure the fairness of a face recognition system, a set of corresponding race or race-related grouping labels $S$ is also specified, $S=\{s_1,s_2,..,s_N\}$. 
Therefore any face dataset can be formed as $D=\{X, Y, S\}$ where $X$ denotes the set of images, $Y$ denotes the set of subject labels, and $S$ denotes the set of sensitive race or race-related labels. Furthermore, a mapping function $f$ plays a significant role in face recognition systems as it maps any given image $x$ into the feature embedding vector $z$. $f$ is selected from a function space $\Omega$ via a loss function $\mathcal{L}$ which measures the performance of a given training set, $D_{train}$, for any of the aforementioned face recognition tasks. Typically, a softmax loss is adopted by state-of-the-art face recognition methods \cite{liu2017sphereface,wang2018additive,deng2019arcface,wang2018cosface} in order to disentangle the feature representation of individual identities within contemporary training datasets. The inductive representation learning is hence a minimisation of the loss function  $\mathcal{L}_{softmax}$, which can be formalised as follows:
\vspace{-.1cm}
\begin{equation}
\quad f^* = argmin(\mathcal{L}_{softmax}(f)),\quad { f \in \Omega} \quad \text{where} \quad 
\mathcal{L}_{softmax} = - \frac{1}{N} \sum_{i=1}^{N} \log \frac{e^{W_{y_i}^{T} z_i + b_{y_i}}}{\sum_{j=1}^{n}{e^{W_{j}^{T} z_i + b_j}}}
\label{eq:softmax}
\end{equation} where $z_i$ is the feature representation of the image $x_i\in \mathbb{R}^{u\times v \times 3}$, $u$ is the weight and $v$ is the height of the $x_i$, within $D_{train}$ belonging to subject class ${y}_i$ and the number of samples is $N$ labelled with $n$ classes. $W_j$ is the $j^{th}$ column of the weights, $b_j$ is the $j^{th}$ column of the bias term, and $d$ is the number of neurons in the last fully-connected layer which is mostly 512. 
Weights and bias term dimensions are $W_j \in \mathbb{R}^{dxn}$ and $b_j \in \mathbb{R}^n$, respectively. Moreover, the selected $f^*$ 
compresses the intra-class distance and expands the inter-class distance between feature embeddings belonging to the same or different subject identity, respectively. Generally, $f$ provides superior approximation over the statistically most predominant population subset within training set, $D_{train}$, such that  $\mathcal{L}_{softmax}$ is minimised.


Additionally, evaluation metrics can quantify how well the selected $f^*$ performs on $D_{test}$. The most common evaluation metric in face recognition, \textit{accuracy}, relates to the probability of correctly predicting the subject label of a face image as $P(y_{\alpha}=\hat{y}_{\alpha})$. Accuracy can be defined as follows,
\vspace{-0.1cm}
\begin{equation}
\text{Accuracy} = \frac{TP+TN}{TP+TN+FP+FN}
\label{eq:acc}
\end{equation} where \textit{true positive (TP)} is the number of the $f^*$ correctly predicts the positive subject label and \textit{true negative (TN)} is the number of the $f^*$ correctly predicts the negative subject label. In contrast, \textit{false positive (FP)} is the number of the $f^*$ incorrectly predicts the positive subject label, and \textit{false negative (FN)} is the number of the $f^*$ incorrectly predicts the negative subject label. Accuracy measures the consistency between predictions and their ground truth values. In a similar vein, the \textit{True Match Rate (TMR)} estimates the number of correct positive predictions made from all possible positive predictions. For instance, a binary face verification task aims to classify whether an image pair $(x_{\alpha},x_{\beta}) \quad where \quad x_{\alpha}, x_{\beta} \in D_{test}$ belongs to the same subject label or not. During testing, the selected $f^*$ predicts the feature representation vectors $z_{\alpha},z_{\beta}$ for the corresponding images $x_{\alpha},x_{\beta}$, respectively. Given images are validated as \textit{"match"} if the similarity between two feature vectors (i.e. \textit{cosine similarity}, $cos(z_{\alpha},z_{\beta}) = \frac{z_{\alpha} \cdot z_{\beta}} {\|z_{\alpha}\| \|z_{\beta}\|}$ ) is greater than a given threshold parameter $threshold$, otherwise as \textit{"non-match"}. \textit{TMR} is the ratio of correctly verified match pairs (two different images from the same subject) over the total number of match pairs. However, neither \textit{Accuracy} nor \textit{TMR} is indicative of failure samples. To investigate such samples, the False Match Rate (FMR) measures how many incorrect non-match or negative predictions $f^*$ are made via feature representation vectors. Furthermore, the False Non-Match Rate (FNMR) refers to the probability of samples of the same subject identity is incorrectly matched. All terms, TMR, TNMR, FMR, and FNMR, can be formalised as follows:

\begin{equation}
\label{eq:mr}
\text{TMR} = \frac{TP}{TP+FN},\quad
\text{TNMR} = \frac{TN}{TP+FN},\quad
\text{FMR}
=\frac{FP}{FP+TN},\quad
\text{FNMR} =\frac{FN}{FP+TN}
\end{equation} Another facial recognition metric, the \textit{ROC curve}, plots TMR against FMR at different thresholds. Lowering the $threshold$ verifies more items as matched, resulting in an increased \textit{FMR} and \textit{TMR}. Furthermore, the racial bias literature commonly measures the variation in performance, indicated by \textit{accuracy} or \textit{FMR}, among racial groups to highlight disparities within each group. However, calculating this deviation varies across studies, as different definitions of standard deviation are used (i.e. sample, population). In this study, we utilise the sample standard deviation for further analysis.

To this extent, we briefly described the selection process of $f$ using the loss function and evaluation metrics of face recognition. Whilst, loss functions help to understand the behaviour of $f$ on $D_{train}$, evaluation metrics help to measure how well the selected $f^*$ maps $D_{test}$ into feature embedding representation space. Consequently, statistical fairness criteria can be considered as a formal property of face recognition systems, including mapping function $f^*$, training $D_{train}$ and test datasets $D_{test}$.
Accordingly, we give the four most commonly used fairness definitions from \cite{kusner2017counterfactual} that are commonplace within racial bias for face recognition.  
 
\noindent
\textbf{Definition 1: \textit{Fairness Through Unawareness}} requires that a machine learning algorithm have an independent conditional probability $P$ of the output given $X$ from $S$ (racial labels). Subsequently, unawareness criteria can be formalised as $P(Y | X) = P(Y | X, S)$. However, removing dependency is impossible for face recognition algorithms due to the high mutual information between facial and racial features. Even though racial labels are not explicitly introduced to the machine learning algorithm, they will implicitly be used in the face representation (algorithm training) via the face images.

\noindent
\textbf{Definition 2: \textit{Individual Fairness}} refers to treating similar individuals coequally, meaning that an algorithm is fair if it gives similar predictions to similar individuals. In order to estimate such criteria, two distance metrics are defined by Dwork \cite{Dwork2012}. These are distance metrics that measure the degree of similarity between individual subjects and measure the difference in the associated prediction outcome between those individual subjects. It can be formalised in face recognition context as if image samples $x_{\alpha}$ and $x_{\beta}$ are similar under a given distance metric $d(x_{\alpha}, x_{\beta})$ depending on $s_{\alpha}, s_{\beta}$ then predictions should be similar $\hat{y}_{\alpha} \approx \hat{y}_{\beta}$ where $\hat{y}_{\alpha}$ and $\hat{y}_{\beta}$ are the predicted labels from corresponding images $x_{\alpha},x_{\beta}$ and $s_{\alpha}, s_{\beta}$ are the sensitive race labels respectively. However, \cite{individual_fairness} discusses how individual fairness is inadequate for ensuring fairness on the grounds of four differing arguments, spanning the insufficiency of similar treatment, systematic bias and arbiters, prior moral judgements, and incommensurability (see \cite{individual_fairness} for a more detailed discussion). 

\noindent
\textbf{Definition 3: \textit{Group fairness (or Statistical parity / Demographic parity})} enforces the predicted subject labels $\hat{Y}$ to be independent of $S$ which can be denoted $P(\hat{Y}|S = s)=P( \hat{Y}| S = s), s \in \{0,1,..r\}$ where $r$ is the number of different sensitive race labels in the set.
Racial bias literature within the face recognition mostly approaches the problem from a supervised machine learning paradigm by considering it as an \textit{group fairness criteria (demographic parity)} \cite{Dwork2012}, which can be satisfied if the race or race-related intersectional groups perform similarly to each other. Unfortunately, such criteria may not ensure fairness as it heavily relies on equalising the acceptance match percentages even though there is little or no training data available for a given racial grouping category within $D_{train}$ \cite{hardt2016equality}. 

\noindent
\textbf{Definition 4: \textit{Equal Opportunity, (or Equalised Odds)}} is satisfied if an algorithm predictions $\hat{Y}$ is independent of $ S$ conditioned on $Y$.
If the criteria is defined for binary categories \cite{hardt2016equality}, it can be denoted $P(\hat{Y} =1 | S = 0, Y=y) = P(\hat{Y} = 1 | S = 1, Y = y), y \in \{0,1\}$. Subsequently, it is adopted by \cite{woodworth2017learning} to multiple class labels.
More simply, the constraint requires that any sensitive race label has equal true positive rates and false positive rates across the other sensitive race labels. It also enforces that the \textit{accuracy} is equally high in all sensitive labels, penalising algorithms that perform well solely on the statistically most predominant such labels. Furthermore, \cite{hardt2016equality} discusses how demographic parity is crippled in the typical scenario in which the target variable $Y$ is correlated with only $S$. On the other hand, \textit{equalised odds} aims to achieve accurate prediction while ensuring predictions are fair concerning a specified sensitive labels, $S$.

As aforementioned, the literature has mainly used \textit{statistical parity or group fairness criteria} to minimise the variation of \textit{accuracy} or \textit{FMR} across sensitive racial groupings labels on datasets.    However, such an aim brings a high dependence on sensitive attributes to be used in fairness criteria above, which may actually increase discrimination \cite{kusner2017counterfactual}. Moreover, little attention has been given to how the sensitive attribute labels, $S$, are assigned, with regard to the potential for bias in the assignment (i.e. labelling) process, and what that potentially means normative \textit{"unbiased"} presumptions for face recognition system design. In the next section, we address these questions by focusing on race and race-related groupings and their conceptualisation.

%% file: tex/3-TowardsRacialGroupFairness.tex
\input{tex/3-TowardsRacialGroupFairness/0-intro.tex}
\input{tex/3-TowardsRacialGroupFairness/1-race.tex}
\input{tex/3-TowardsRacialGroupFairness/2-skin-tone.tex}

\input{tex/3-TowardsRacialGroupFairness/3-facial-phenotypes.tex}

\input{tex/3-TowardsRacialGroupFairness/4-conclusions.tex}

%% file: tex/3-TowardsRacialGroupFairness/0-intro.tex
\vspace{-.4cm}
\section{Towards Racial Group Fairness}
\label{sec:3}

Most studies on racial bias within face recognition, with a few exceptions \cite{sun2021nfw,xue2020auditing}, use the criteria of \textit{group fairness (demographic parity)} to evaluate and mitigate both data and algorithmic bias. However, \textit{group fairness criteria} relies on sensitive attribute labels such as race, ethnicity or skin tone and uses performance evaluation metrics such as accuracy or false match rate. Subsequently, stratification of the complex and multi-faceted concept of race into abstract race-related categories becomes necessary in order to address racial bias \(\textit{group fairness}\) as the categories allow us to assess whether the final performance of a given face recognition system is fair and satisfy (\textit{group fairness criteria}). Accordingly, the face recognition literature mainly utilises either race (e.g. African, Asian, etc.) or race-related grouping categories (e.g. skin tones, facial phenotypes etc.). However, with regard to racial stratification, this construction of race or race-related groupings also brings with its and additional set of challenges. For example, early attempts at the conceptualisation of race itself inherited racial bias, as the way race is defined and understood is influenced by preexisting prejudices and discriminatory beliefs \cite{benjamin2019race,zuberi2001thicker}. As a result, the way race is conceptualised may perpetuate and reinforce existing forms of racial inequality \cite{zuberi2001thicker}. Moreover, exposing or using such racial origin identifies the representation of a particular group and may lead to potential racial profiling and associated inequality \cite{mozur2019one}.
Additionally, race or skin tone grouping strategies can limit the scope of any study as they fail to capture the whole aspect of the racial bias problem within face recognition where it needs to consider both multi-racial or less stereotypical members of such groupings \cite{buolamwini2018gender,mitchell2020diversity}. Hanna \cite{hanna2020towards} discussed treating race as an attribute rather than a structural, institutional, and relational phenomenon and ignoring its multidimensional factors can result in missing important aspects of algorithmic fairness. Finally, many researchers do not provide detailed background about their racial categorisation process \cite{scheuerman2020we}, which makes such race-related groupings even more insurmountable in effectively addressing racial bias. Published datasets and related research work rarely contain details about how racial groups are determined or how racial bias evaluation metrics are designed \cite{scheuerman2020we}. In addition to the aforementioned points, many studies \cite{raji2020saving,benthall2019racial,scheuerman2020we} highlight the potential risks of omitting the details of the racial categorisation strategy along with the appropriate context for use.

In this section, we delve into the racial bias (\textit{group fairness criteria}) within face recognition. We examine how race and race-related grouping categories are constructed, the significance of accurately defining these categories and the potential risks and consequences of using and evaluating them in face recognition systems. We classify groupings under the three most predominantly used categories: race, skin tone, and facial phenotype. We discuss the grouping strategies in each category together with their potential positive and negative impact and describe the details of subcategories where they have been used. Furthermore, we cover the literature on annotation processes of grouping categories and summarise recent literature along with face datasets by organising them under their grouping strategies in Table \ref{tab:frdatasets}.




%% file: tex/3-TowardsRacialGroupFairness/1-race.tex
\vspace{-.4cm}
\subsection{Race}
\label{sec:3:1}
Race, as a term for human categorisation based on varying factors, is a controversial concept related to sociology, psychology, biology, ethnology, and cultural anthropology, whose definition varies across different fields and throughout history. Within biology, for example, the race concept has been differentiated into three different kinds: genetic, morphological and psychological, which are all widely disputed \cite{quine1953three}. Race was first delineated by European naturalists and anthropologists to establish population-based research on human diversity \cite{muller2014race}. In the seminal early scientific work of 1758, Systema Naturae \cite{linnaeus1758systema}, Carl Linnaeus categorises humans into four different groups: \textit{European white, Americanus rubescens (American reddish), Asiaticus fuscus (Asian tawny), Africanus niger (African black)} using a combination of continental (geographic) and observational (skin tone) terminology. Subsequently, several attempts were made to classify and group humankind in such a manner in order to use it in societal statistics \cite{zuberi2001thicker,zuberi2008white,hanna2020towards}. Most of the work was problematic (by the standards of today) or error-prone (even by the standards \textit{of the day}) as it reflected the biased ideologies of researchers, politicians and institutions of that time \cite{zuberi2001thicker}. However, such definitions and classifications were adopted by the national census infrastructure across many jurisdictions  \cite{hanna2020towards}. The work of Khalid Muhammad \cite{muhammad2019condemnation} reveals how anecdotal, hereditarian and pseudo-biological race theories transformed into statistics and social surveys. Furthermore, Zuberi \cite{zuberi2001thicker} addresses the complicated history of racial stratification and its evident impact on social and natural sciences. Consequently, he defines race as a biological notion of physical difference grounded in an ideology \cite{zuberi2001thicker}. 

Within face recognition, subject face images form the primary information source that encapsulates these race-related biological and physical differences, which are then combined with additional information, including gender, age, pose, facial expression and contextual aspects such as scene background, illumination, subject clothing and facial accessories such as glasses, facial hair, jewellery and makeup. On this basis, it becomes possible to adopt any such ideology via the use of racial groupings and classifications that are introduced to face recognition with the aim of quantifying racial bias. However, despite this potential, an increasing number of face recognition studies instead adopt different variations of racial categorisation \cite{grother2019face,robinson2020face} without any reference to the underlying critical theory of such categorisation and how they are defined \cite{zuberi2001thicker,zuberi2008white,hanna2020towards}. More worryingly, racial annotation of face imagery has now become the initial step in many proposed face recognition approaches aiming to address racial bias, but the crucial decision-making on how and why a given racial categorisation is defined remains subjective, arbitrary and largely undocumented \cite{miceli2020between}.

\input{tables/frdatasets}

Previously, racial categories made an initial appearance within automated facial analysis via the task of race classification. For example, \cite{Yongsheng2005} propose feature extraction-based techniques for race classification using the MORPH \cite{ricanek2006morph}, and FERET datasets \cite{phillips2000feret} to predict ${\textit{Caucasian, South Asian, East Asian, and African}}$ racial classification. Later studies \cite{ricanek2006morph} extend the MORPH dataset for face recognition and analysis tasks (identification, recognition, and verification) by providing additional ground truth labels spanning age, gender, race, height, weight, and eye position. Subsequently, DCNN-based methods were introduced for race classification \cite{Greco2020,samir2020, Ahmed2020}. The work of \cite{Greco2020} proposes the large-scale VGGFace2 Mivia Ethnicity Recognition (VMER) dataset, composed of more than 3 million face images annotated with four ethnicity categories, namely ${\textit{African American, East Asian, Caucasian Latin and Asian Indian}}$, and provides comprehensive performance analysis for several contemporary deep network architectures, namely VGG-16, VGG-Face, ResNet-50 and MobileNet v2. Although such race classification techniques are not necessarily used as a proxy for facial image annotations with regard to the study of racial bias within face recognition, these public datasets containing race labels and their associated racial groupings are widely adopted \textit{de facto} by the face recognition research community. As we illustrate in Table \ref{tab:frdatasets}, the most commonplace face recognition datasets containing race labels \cite{wang2019racial, buolamwini2018gender} use three grouping strategies, namely race, skin tone and facial phenotypes. Similar to race classification, broader racial groupings such as \textit{\{African, Asian, Indian and Caucasian\}} or binary racial groupings such as \textit{\{Black, White\}} are also commonly followed by many datasets creators \cite{wang2019racial,buolamwini2018gender}.  

Recently, the most commonly used face recognition evaluation dataset, a subset of MS-Celeb-1M \cite{guo2016ms} released as the RFW dataset \cite{wang2019racial}, was constructed to measure relative face verification performance across four different racial groupings: \textit{\{African, Asian, Indian, Caucasian\}}. FairFace \cite{karkkainen2021fairface} is another dataset, again drawn as a subset from the larger YFCC-100M Flickr dataset \cite{thomee2016yfcc100m}, which supplements this earlier set of four labels with two additional racial groupings, \textit{\{Middle East, Latino\}} to evaluate racial bias more broadly. In addition, UTKFace \cite{zhifei2017cvpr} is a large-scale face dataset with five different racial groupings, namely \textit{\{Asian, Black, Indian, White and
Others (like Hispanic, Latino, Middle Eastern)\}}, for various tasks spanning face detection, age estimation, and age progression/regression. This variation in racial groupings illustrated more extensively in Table \ref{tab:frdatasets}, highlights the ambiguity and uncertainty behind the race concept upon which the presence of bias is ultimately being evaluated. Consequently, this inconsistency of racial groupings, its historical and geographic instability within the face recognition research literature and the commonplace adoption of ill-defined race concepts that are littered with a problematic history with social statistical science make effective performance evaluation and quantification very challenging within the racial bias problem space. 

Similarly, Khan \cite{khan2021one} identify four specific problems with the racial categories: (1) categories are not clearly defined and are often loosely associated with geographic origin, (2) categories that are extremely broad, with continent-spanning construction that results in individuals with vastly different physical appearance and ethnic backgrounds being grouped incongruously into the same racial category, (3) categories narrow down the differences between ethnic groups with distinct languages, cultures, separation in space and time, and phenotype into the same racial category. (4) assigning a single racial category to a face example for performance evaluation of any form of automated analysis, including face recognition, is not an ideal solution as it cannot capture a substantial proportion of the distribution of diversity and variation within the human race.

In parallel with Khan, Raji \cite{raji2020saving} discusses three ethical tensions when auditing commercial facial processing systems, where there exists a requirement to annotate face imagery with race or race-related categories. \textit{Privacy and Representation:} Collecting a diverse and representative dataset for facial recognition can bring privacy risks for individuals included in the dataset. Furthermore, potential consent violations may arise during the data collection process, for example, for the IBM Diversity in Faces dataset \cite{merler2019diversity}, which was sourced from images on the public image-sharing platform Flickr that were uploaded under very permissive licensing terms (Creative Commons). However, it later emerged that the individuals within the photos did not necessarily consent to be included within the face recognition dataset \cite{solon2019facial}. \textit{Intersectionality and Group-Based Fairness:} Intersectionality is based on the idea that the experience of an individual cannot be fully understood by looking at one aspect of their identity. However, when evaluating group fairness in facial recognition systems, assigning individuals to a racial category and performing disaggregated analysis to account for multiple categories is often necessary. This type of analysis can help to identify and address potential biases, but it may not fully capture how varying components of a face recognition processing pipeline interact to recognise individual features across individuals with multiple marginalised identities. \textit{Transparency and Overexposure:}  Although sharing details of the dataset development process and publicly disclosing named audit targets can help to clarify the scope of the audit and the context in which results should be interpreted. This can also result in targeted over-fitting (i.e. \textit{"cheating"}) in order to optimise system performance on the audit. Moreover, this can also lead to pressure to make the audit more operationally relevant to real-world deployment. For example, some institutions have removed or restricted access to their facial recognition benchmark assets following their inclusion in audits, which can compromise the performance validation of future systems and make it more expensive and difficult for other researchers to evaluate relative performance changes in the field \cite{raji2019actionable}.

Finally, although many more studies discuss the possible negative consequences of using racial categories in face recognition datasets,  Table \ref{tab:frdatasets} proves that such racial categories have become commonly used and increasingly contributed within the literature. The lack of work on alternative race-related grouping strategies or fairness criteria that do not rely on any racial category forces racial bias literature to address racial bias using such commonly defined racial categories. Considering the problems that arise with racial categorisation, the current status of research that uses racial categories (\textit{still}) does not paint an optimistic picture of the global face recognition research community collaboratively tackling racial bias. As information of racial or ethnic origin remains sensitive \cite{hepple2010new}, from these observations across the face recognition field, we agree with the findings of several major studies \cite{zuberi2001thicker,maddox2018racial,mozur2019one,benthall2019racial,raji2020saving,krishnapriya2020issues, mitchell2020diversity,muhammad2019condemnation} that already highlight the adverse effects of the use of racial categories and their suggestion that researchers should either avoid revealing such sensitive data or provide an appropriate context for use. Furthermore, transparent provision of the ethical considerations together with any details of the racial annotation process in use and the intended possible use cases, limitations, and risks of the designed solution, should be made by the originating team in all cases \cite{gebru2021datasheets}. 

%% file: tables/frdatasets.tex
\begin{table}[htp]
  \rowcolors{1}{rowzebra}{white}
  \resizebox{\columnwidth}{!}{%
  \begin{tabular}[t]{llllll}
  \toprule
    \rowcolor {white}
    \textbf{Dataset Name}  & \textbf{Year} & \textbf{Grouping Categories} & \textbf{Images} & \textbf{Source} \\ 
    
    \midrule
    \multicolumn{5}{c}{\textbf{\textbf{Race}}}  \\
    \hline
     
    ColorFERET \cite{phillips2000feret} & 1993 & White, Asian, Black, Others & 14K & Participants' photographs \\
    MORPH   \cite{ricanek2006morph} & 2006 & Caucasian, Hispanic, Asian, or African American & 55K & Public Records \\
    UTK Face \cite{zhifei2017cvpr} & 2017 & \begin{tabular}[t]{@{}l@{}}Asian, Black, Indian,  White and \\ Others (Hispanic, Latino, Middle Eastern)\end{tabular} & 20K & \begin{tabular}[t]{@{}l@{}} MORPH, CACD,\\ online resources \end{tabular}\\ 
    IJB-C \cite{maze2018iarpa}& 2018 & \begin{tabular}[t]{@{}l@{}}North American, South America, Western Europe, \\ South West Africa, East Europe, East Africa-Middle \\ East, South East Asia, India, China, East Asia\end{tabular} & 31K & \begin{tabular}[t]{@{}l@{}} Public, law enforcement \\ databases, social media \end{tabular} \\
    RFW \cite{wang2019racial} & 2019 & African, Asian, Caucasian, Indian & 45K & MS-Celeb \cite{guo2016ms} \\
    DemogPairs  \cite{hupont2019demogpairs} & 2019 & Asian, Black, White& 10.8K & CWF, VGGFace1-2 \cite{yi2014learning,parkhi2015deep, cao2018vggface2}  \\
    BUPT-Balanced \cite{wang2020mitigating} & 2020 & African, Asian, Caucasian, Indian & 1.3M & MS-Celeb \cite{guo2016ms} \\
    VGGFace2 1200   \cite{yucer2020exploring} & 2020 & African, Asian, Caucasian, Indian & 1M & VGGFace2 \cite{cao2018vggface2} \\
    FairFace   \cite{karkkainen2021fairface}& 2021 & \begin{tabular}[t]{@{}l@{}}Black, East Asian, Indian, Latino, Middle Eastern, \\ Southeast Asian, and White\end{tabular} & 108K & \begin{tabular}[t]{@{}l@{}}Flickr, Twitter, newspapers,\\online resources\end{tabular} \\
    CASIA-Face-Africa   \cite{muhammad2021casia}  & 2021 & Hause (Sudan, Chad, Binin, Ivory Coast), Non-Hause & 38K  & Subjects from  Nigeria \\
    DiveFace  \cite{sensitivenets} & 2021 & \begin{tabular}[t]{@{}l@{}l@{}} (Japan, China, Korea),  (Europe, North America, and Latin \\America) (Sub-Saharan Africa, India, Bangladesh, Bhutan)    \end{tabular} & 120K & MegaFace \cite{kemelmacher2016megaface} \\
    
    \midrule
    \multicolumn{5}{c}{\textbf{\textbf{Skin}}~ \textbf{\textbf{Colour}}} \\
    \hline
  
    IJB-B   \cite{whitelam2017iarpa} & 2017 & \begin{tabular}[t]{@{}l@{}}1-6 skin tones (increasing in darkness)\end{tabular} & 1K & \begin{tabular}[t]{@{}l@{}} 1M FreeBase Celebrity List\end{tabular} \\
    PPB  \cite{buolamwini2018gender} & 2018 & Light, Dark skin tones (Fitzpatrick I-III,IV-VI) & 68K & Gov. Official Profiles \\
    Fair Face Challenge   \cite{sixta2020fairface} & 2020 & Light, Dark skin tones (Fitzpatrick I-III,IV-VI) & 152K & \begin{tabular}[t]{@{}l@{}}Flickr, Twitter, newspapers,\\online resources\end{tabular} \\
    Casual Conversations   \cite{hazirbas2021towards}& 2021 & Fitzpatrick Skin Tones & 45K* & Vendor data \\
    Globalface-8  \cite{wang2021meta} & 2021 & \begin{tabular}[t]{@{}l@{}}ITA base 8 skin tones (Tone I-VIII)\end{tabular} & 2M & \begin{tabular}[t]{@{}l@{}} 1M FreeBase Celebrity List\end{tabular} \\
    Balancedface-8   \cite{wang2021meta} & 2021 & \begin{tabular}[t]{@{}l@{}}ITA base 8 skin tones (Tone I- VIII)\end{tabular} & 1.3M & \begin{tabular}[t]{@{}l@{}} 1M FreeBase Celebrity List\end{tabular} \\
    IDS-8   \cite{wang2021meta} & 2021 & \begin{tabular}[t]{@{}l@{}}ITA base 8 skin tones (Tone I-VIII)\end{tabular} & 10K & \begin{tabular}[t]{@{}l@{}} 1M FreeBase Celebrity List\end{tabular} \\ 
    
    \midrule
    \multicolumn{5}{c}{\textbf{\textbf{Facial}}~\textbf{\textbf{Phenotypes}}} \\
    \hline
  
    Diversity in Faces  \cite{merler2019diversity}& 2019 & \begin{tabular}[t]{@{}l@{}}ITA 6 skin tone, Craniofacial distance, area,\\  ratio, Facial region contrast\end{tabular} & 0.97M & YFCC-100M \\
    VGGFace2 \cite{cao2018vggface2} - \cite{yucer2022measuring}& 2018 & \begin{tabular}[t]{@{}l@{}}Fitzpatrick Skin Tones, Nose Shape, \\ Eye Shape, Mouth Shape, Hair Type\end{tabular} & 3.3M & Google Image Search \\
    RFW \cite{wang2019racial} -\cite{yucer2022measuring}& 2019 & \begin{tabular}[t]{@{}l@{}}Fitzpatrick Skin Tones, Nose Shape, \\ Eye Shape, Mouth Shape, Hair Type\end{tabular} & 45K & MS-Celeb \cite{guo2016ms} \\ 
    
    \bottomrule
  \end{tabular}%
  }
  \caption{Overview of most prominent face recognition datasets categorised by racial groupings, including dataset size and image sources.}
  \Description{}
  \label{tab:frdatasets}
  \vspace{-0.9cm}
  \end{table}

%% file: tex/3-TowardsRacialGroupFairness/2-skin-tone.tex
\vspace{-.4cm}
\subsection{Skin Tone}
\label{sec:3:2}
\begin{figure}[!t]
  \centering
  \includegraphics[width=\linewidth]{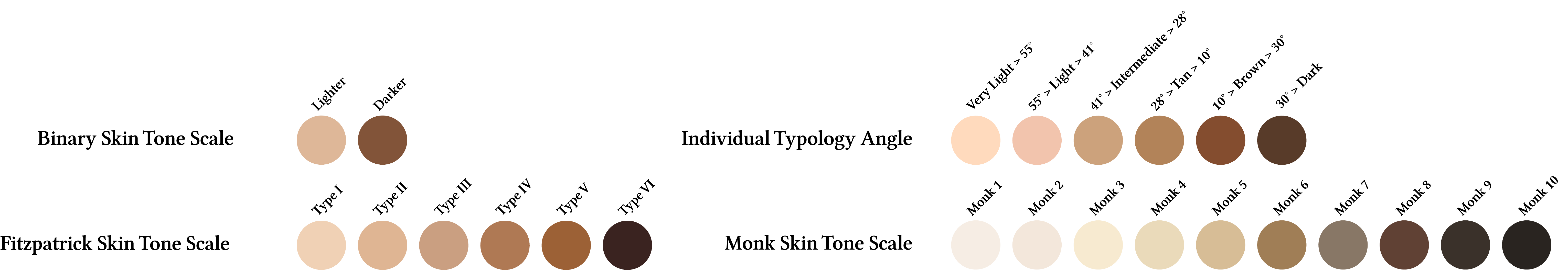}
  \caption{ Four different skin tone scales used for racial bias analysis within the context of face recognition.}
  \label{fig:skincolors}
  \vspace{-0.4cm}
\end{figure}

Human skin tone ranges can vary from saturated black to off-white pale, representing one of the key race-characterising traits. Variations in skin tone among humans have been traditionally used to classify people into race or race-colour identities \cite{harris1993whites} as skin tone variation caused by genetic differences  (also exposure to the sun). In terms of its biological foundation, melanin is a group of natural pigments many organisms produce that is the fundamental driver for skin tone variance. The outermost layer of skin, the epidermis, contains melanin pigments, including red/yellow phaeomelanin and brown/black eumelanin \cite{muehlenbein_2010}. Moreover, haemoglobin, beta carotene and bilirubin in the second layer of the skin, the dermis, absorb light and contribute to the overall pale/yellow/orange tint of human skin. It is primarily the quantity and type of melanin present in the skin that determines the skin tone. Genetic and hormonal factors control the production process of melanin, while direct exposure to ultraviolet radiation (UVR) accelerates its production \cite{fitzpatrick1961melanin} (hence the common phenomena of that a temporary darkening of skin tone is associated with exposure to sunlight - hence the expression \textit{"sun tan"} where specifically the use of the term \textit{tan} is indicative of a darkening in colour). 

Over the past centuries, methods for categorising skin tone have evolved from verbal race-related descriptions (that would potentially be seen as derogatory today) with skin colour categories as \textit{"white", "yellow", "black", "brown"}, and \textit{"red"} \cite{oliver2017race}, and colour-matching-based methods such as the Von Luschan scale \cite{von1897beitrage}. The Von Luschan scale \cite{von1897beitrage} uses 36 coloured glass tiles for skin colour comparison and was commonly used to racially categorise the population until the mid-20th century. Later, the Fitzpatrick Scale, established in 1975, became the most commonly used skin tone scale in dermatology and medicine. Accordingly, a plethora of work on racial bias within face recognition adopts the Fitzpatrick scale to measure and mitigate racial bias. However, Fitzpatrick's skin tone measurement was initially designed based on a subjective self-reporting or dermatology expert assessment which are often inconsistent and unreliable \cite{Howard21}. Subsequently, reflectance spectrophotometry and colourimetry methods \cite{ly2020research} have now become preferential in medical skin tone assessment over earlier methods due to increased accuracy and consistency. Whilst colourimeters quantify the appearance of a tone on the skin, a spectrophotometer measures the spectral characteristics of the skin colour. Such devices convert light reflectance data from the skin into colourimetric values for estimating chromophores in the skin \cite{Munidasa2018}. Subsequently, Individual Typology Angle (ITA) \cite{chardon1991skin} has been proposed to classify human skin colour using spectrophotometric measurements. More recently, a new extended skin colour scale, the Monk Skin Tone (MST) scale, was proposed with ten different tones to use in computer vision applications \cite{sgoogle}, whilst the global cosmetics industry has traditionally used a separate set of scales \cite{caisey2006skin}. The contrasting examples of these various skin tone scales are illustrated in Fig. \ref{fig:skincolors} where we can see a sharp contrast between categorisation in binary, Fitzpatrick, ITA or MST skin tone groupings. However, skin tone scale grouping strategies alone carry various concerns for the mitigation of racial bias within face recognition. We discuss these concerns under three divisions as follows:

\noindent
\textit{Erroneous Skin Tone Annotation:} Firstly, most skin tone scales are designed to measure skin tone on physical human subjects in a medical or dermatological context. By contrast, face recognition systems instead used such annotations for digitally captured face images that form part of the training and test data sets (see Section \ref{sec:4:1}). Moreover, such face image samples are commonly yielded from public domain sources (i.e. internet search engine-based image retrieval - \textit{"in-the-wild"}), and as such, this uncontrolled imagery exhibits enormous variation in both environmental and subject conditions at the point of image capture. Similarly, \cite{Kolkur2017HumanSD} summarises such varying conditions that affect skin-colour detection in the visible spectrum as scene illumination, camera characteristics, demographic characteristics (race, age, gender), and other factors (make-up, wearing glass, hairstyle, head pose). Such varying factors make effective skin tone annotation challenging and result in erroneous skin tone assignment for given subjects/samples. Furthermore, human annotators often bring subjectivity and inconsistency to the resulting annotation labels far more so than other image labelling tasks (c.f object/scene categorisation), whereas skin tone annotation ideally needs to be objective, consistent, and repeatable \cite{miceli2020between}. Specifically, \cite{krishnapriya2021analysis} highlights the uncertainty within the human-based categorisation of skin tones from digital image and proposes the use of automated skin tone assignment as a means of potentially achieving speed, scalability and consistency. However, the consistent skin tone annotation of a given subject under the aforementioned image variations remains a pertinent issue with such automated solutions - one that in itself presents a circular occurrence of bias within facial processing.

\noindent
\textit{Narrow Representation of Scales:} Secondly, the most commonly used skin tone scales used for accessing aspects of racial bias are either too narrow in terms of their discretisation of the skin tone spectrum (e.g. Binary Skins Groups, Fig. \ref{fig:skincolors}) to facilitate capture of the foundational reasons for bias or alternatively offer the less representative capability for specific groups (e.g. Fitzpatrick Skin Types vs Monk Skin Tone Scale, Fig. \ref{fig:skincolors}) \cite{Howard21}.

\noindent
\textit{Skin Tone as a Single Dimension of Race:} Thirdly, race is a multi-faceted concept conflating other phenotypic facial traits such as lips, eyes, hair and face shape. Solely aligning racial grouping with skin tone only transforms the racial bias problem into a single-faceted problem. Moreover, there is no clear evidence that skin tone alone is the primary driver for disparate false match rates within face recognition performance \cite{krishnapriya2020issues}. Accordingly, several studies suggest considering other race-related facial attributes, including lips, eye, and face shape when measuring racial bias in this context \cite{muthukumar2018understanding,muthukumar2019color} in order to enable improved interpretation and derivation of bias factors. Accordingly, a consensus is beginning to emerge on skin tone assignment and the appropriate quantification of skin tone within digital facial images as used in face recognition research. Various studies \cite{buolamwini2018gender,sixta2020fairface,hazirbas2021towards} measure the racial bias in face recognition using either binary skin groupings, the Fitzpatrick Skin Types \cite{fitzpatrick1988validity}, or ITA \cite{chardon1991skin} as depicted in Figure \ref{fig:skincolors}.

\vspace{-.3cm}
\subsubsection{\textbf{Binary Skin Tone Scale}}

The first research study that is based on the usage of binary skin/race groupings appeared in sociological research on race and race relations \cite{omi2014racial}. Focusing on white-black race relations in the United States brings expensive socio-economic data and analysis around such binary groupings \cite{gonzalez2019exploring}. Accordingly, the adaption of binary skin/racial groupings into computer vision tasks such as skin tone estimation, race classification and racial bias of face analysis systems started from this simple categorisation viewpoint. In order to model skin colour on imagery, several studies \cite{kakumanu2007survey} proposed quantitative colour-space divisors (i.e. a dark-light pixel colour threshold) and simply grouped skin colours into binary categories. In the racial bias context, many studies adopt such a darker-lighter skin tone grouping by either narrowing the Fitzpatrick scale or dividing subject skin tone variance into binary categories. One of the seminal works in the field, Gender Shades \cite{buolamwini2018gender}, uses darker-lighter skin tone categories on the Pilot Parliament dataset to demonstrate the algorithmic performance disparities in both gender classification and face recognition tasks. Another example is the Fair Face Challenge study \cite{sixta2020fairface}, which suggested researchers used a requantised (narrower) set of Fitzpatrick skin tone categories as per Gender Shades \cite{buolamwini2018gender}. Despite binary skin tone categories are being the most straightforward grouping strategy in terms of automatic image annotation, in practice, it often obscures the complexity of race concept and results in the mis-quantification of the racial bias problem across solutions where the ultimate aim is unbiased performance across any skin tone variant. This is attributable to imaging effects such as skin reflectance, which was shown by Cook \cite{cook2019demographic} to have a very significant net effect on the average biometric performance when considered across three different skin reflectance groupings within face recognition. As such, the use of simple binary groupings is known to result in erroneous or conflicting group interpretations, whilst broader groupings such as Fitzpatrick Skin Types claim to be more robust against this issue \cite{yucer2022measuring}.

\vspace{-.3cm}
\subsubsection{\textbf{Fitzpatrick Skin Tone Scale}}
The dermatologist Thomas B. Fitzpatrick developed his Fitzpatrick Skin Tone Scale to assess the propensity of the skin to burn during photo-therapy (i.e. the treatment of skin conditions using intense ultra-violet light sources). Initially, four different types ranging from Type I (always burns, does not tan) to Type IV (rarely burns, tans with ease) were released by \cite{fitzpatrick1975soleil}. Later, he extended his scale to include a broader range of skin types (Type V and IV) \cite{fitzpatrick1988validity} in order to offer a more granular representation across darker skin tones. The widespread adoption of this work within medical research studies \cite{sommers2019fitzpatrick,pichon2010measuring} subsequently influenced early computer vision research studies considering skin tone. Within the racial bias literature, the Gender Shades study \cite{buolamwini2018gender} was the first to gather attention around the use of the Fitzpatrick Skin Tone Scale within an automated facial image analysis context. Subsequent studies then released varying datasets, all using the Fitzpatrick scale on this basis \cite{sixta2020fairface,hazirbas2021towards,yucer2022measuring}. Even recently, the extensive Casual Conversations Dataset \cite{hazirbas2021towards} containing 45K videos makes use of Fitzpatrick skin tone labels for its racial grouping strategy. However, other researchers have raised concerns about using the Fitzpatrick scale on image-based visual tasks \cite{Howard21}. Primarily, the Fitzpatrick scale was not initially designed for image-based skin tone estimation; hence, its evaluation methodology relies on physical skin measurement. As a result, its use can cause inconsistent skin tone assignment when applied on images \cite{krishnapriya2021analysis}. Consequently, \cite{Howard21} observes how challenging it is to robustly assign darker skin tone labels within the Fitzpatrick scale when faced with a significant imaging variance and suggests avoiding the use of such skin tone assignments
ascertained from images captured under uncontrolled or unknown conditions.


\vspace{-.3cm}
\subsubsection{\textbf{Individual Typology Angle } (\textit{ITA})} Another skin tone scale based on a spectrophotometric evaluation of skin colour is introduced by Chardon in 1991 \cite{chardon1991skin}. This method utilises the reflection of skin light via spectrophotometers that measure $LaB$ colour values of the skin ($L$: Lightness. $a$: Red/Green Value. $b$: Blue/Yellow Value) to represent the intensity of pigments such as carotene, haemoglobins, phaeomelanin, and eumelanin. Accordingly, Chardon proposes six physiologically skin categories: \textit{\{very light, light, intermediate, tan, brown, and dark\}} estimated via equation of ITA $ITA = \arctan(\frac{L - 50}{b}) \times \frac{ 180}{\pi}$. \textit{ITA} projects skin colour volume into $LaB$ colour space, and is used to categorise skin angle via the associated \textit{ITA} classification thresholds (see Fig. \ref{fig:skincolors}) \cite{krishnapriya2021analysis}. As the \textit{ITA} solely relies on precise and objective skin tone measurements, it is considered more accurate than traditional visual assessments. Furthermore, it provides a better representation of both the diversity and contributory factors associated with skin tone
\cite{wu2020utilization,kinyanjui2019estimating}. On the other hand, the utilisation of \textit{ITA} scores and categories varies in the literature; Wang \cite{wang2021meta} constructs three large-scale face recognition datasets containing four or eight different skin tone groupings based on \textit{ITA} scores and releases the corresponding skin tone labels for each face image with the datasets. The Diversity in Faces dataset \cite{merler2019diversity} also adapts \textit{ITA} (using six categories) as they find ITA both a more practical and straightforward method for measuring facial skin tone. However, akin to the earlier aforementioned issues with skin tone estimation from digital face images, inconsistent and uncontrolled imaging conditions again impact accurate and reliable \textit{ITA} assessment \cite{wu2020utilization,krishnapriya2021analysis}.

\vspace{-.3cm}
\subsubsection{\textbf{Monk Skin Tone (MST) Scale}}

Most recently, the work of Ellis Monk \cite{sgoogle} produced a new 10-shade skin tone scale designed to facilitate the construction of more representative datasets for the development of on-line consumer services. Although the associated study discusses the aforementioned limitations of prior work on skin tone groupings such as the Fitzpatrick Skin Tone Scale \cite{fitzpatrick1988validity}, it does not provide any detail for the practical application of the new 10-shade scale or any additional guidance via the provision of an exemplar dataset \cite{sgoogle}.

Overall, this section provides an overview of skin tone characterisation approaches and their associated quantification methodologies spanning both digital imagery and physical dermatological examination. Accordingly, we summarise the most common skin tone scales and discuss the challenges of applying such estimation approaches to the skin tone labelling task within face recognition datasets. Furthermore, we outline all of the face recognition datasets in the research literature that use varying skin tone scales in Table \ref{tab:frdatasets}. As skin tone-based groupings become widely used for racial bias evaluation studies, many benchmark datasets are unfortunately annotated with varying skin tone scales and with varying levels of labelling robustness. Although utilising skin tone scales as a labelling concept for face recognition datasets avoids otherwise using sensitive or ill-defined racial categories, the subjectivity of human-based skin tone annotation, the inconsistency of facial image capture conditions and most pertinently the fact that the skin tone is only one dimension of race all make it an imperfect mechanism for the quantification of racial bias within face recognition. As a result, we suggest developing a broader strategy based on the use of high-accuracy, consistent and reliable facial phenotypes that can instead analyse the true relationship between facial features and racial bias. Consequently, we believe such approaches enable investigation across every facial trait and hence bring greater granularity to the quantification of racial bias within face recognition whilst avoiding the use of problematic racial categorisation.

%% file: tex/3-TowardsRacialGroupFairness/3-facial-phenotypes.tex
\vspace{-.4cm}
\subsection{Facial Phenotypes}
\label{sec:3:3}

Human phenotypic variation refers to variation over the set of morphological and observable characteristics of an individual, which is the result of both genetic and environmental factors \cite{guo2014variation}. Such variation is most observable on faces as the face is identified as a \textit{"biological billboard of our identity"} \cite{claes2014toward}. Subsequently, many studies \cite{sesardic2010race,ousley2009understanding} focus on the impact of human phenotype characteristics (such as morphological attributes) on race. For example, the \textit{Shades of Race} study \cite{Feliciano2016} investigates the marginal effects of phenotypic characteristics, including skin tone, lips, nose, hair and body type on racial categorisation. Moreover, Zhuang \cite{zhuang2010facial} considers 21 craniofacial measurements such as face width, length, nose dimensions and eye corner locations in order to show statistically significant differences in facial measurements between four racial grouping, which are \textit{\{Caucasian, Hispanic, African, other (mainly Asian)\}}. Therefore, a race-related facial phenotypes can be considered to be specific to such facial characteristic attributes, which can then also be correlated to race (\textit{"Phennotopically similar individuals are expected to be genetically more similar as well."}, \cite{hopman2020facing}). On the other hand, facial phenotypes such as skin tone or hair colour do not identify racial categories within themselves, but they can combine with other attributes to identify a broader racial grouping \cite{m2020tentacular}. Furthermore, this correlation between such facial phenotypes and racial categories may not be readily visible or clearly delineated, which is in fact highly desirable when we aim to curb the continued use of problematic historical racial categorisation approaches and the disclosure of sensitive racial categories\cite{rothman1998genetic} (see Sec. \ref{sec:3:1}).

Moreover, Maddox \cite{maddox2004perspectives} explains \textit{racial appearance bias} as a negative disposition toward phenotypic variations in facial appearance. He also discusses how race-conscious social policies may fail to address racial bias in the societal treatment and socioeconomic outcomes of disadvantaged groups \cite{maddox2018racial}. For example, many studies show that individuals with more stereotypical racial appearance suffer from poorer socioeconomic outcomes than those with less stereotypical appearance for their race \cite{maddox2018racial,skinner2015looking,kahn2011differentially}. Additionally, the sole use of race or skin tone categories to quantify racial bias is limiting as they do not account for multi-racial individuals or those who exhibit less stereotypical racial traits. Within this context, an improved understanding of the role of phenotype variation may complement existing solutions that attempt to address racial bias \cite{maddox2004perspectives}. 

\input{tables/phenotypes}

A set of race-related facial phenotype attributes such as skin tone, nose shape, and lip shape are of primary interest for quantifying and addressing racial bias in face recognition. Furthermore, the recent work of \cite{terhorst2021comprehensive} show that non-explicit racial attributes (accessories, hairstyles or facial anomalies) conflated with explicit racial attributes (skin tone, nose shape or eye shape) strongly affect recognition performance. This study discusses the need to investigate each attribute in order to achieve robust, fair and explainable face recognition solutions \cite{terhorst2021comprehensive}. Such requirements directly contradict the use of more traditional racial groupings as they remain a high-level, yet impoverished representation to facilitate elaborate performance interpretation \cite{barocas2021designing}.
Subsequently, a plethora of work highlighting the shortcomings of race and skin tone-based categorisation (discussed in Sec. \ref{sec:3:1} and \ref{sec:3:2}) push the current direction of research into phenotype-based categories \cite{yucer2022measuring}. One of the example studies, \textit{Diversity in Faces} \cite{merler2019diversity}, provides a new large-scale facial data that implements annotations across ten facial coding schemes in order to provide human-interpretable quantitative measures of intrinsic facial features. The study comprises an extensive set of facial annotations spanning intrinsic facial features to include craniofacial distances, areas and ratios, symmetry and contrast, skin tone (\textit{ITA}), age, gender, subjective annotations, head pose and image resolution that are listed in Table \ref{tab:phenotypes_a}. However, despite its potential to date this \textit{Diversity in Faces} is not publicly available due to increased sensitivity around subject privacy and consent issues (as discussed in Sec. \ref{sec:3:1}).

In parallel, \cite{yucer2022measuring} proposes a phenotype-based evaluation strategy for racial bias within face recognition. The study categorises representative racial characteristics on the face and explores the impact of each characteristic phenotype attribute: skin tone, eyelid type, nose shape, lips shape, hair colour and hair type. They annotate these attributes for two different publicly available face datasets: VGGFace2 (test set) \cite{cao2018vggface2}, and RFW \cite{wang2019racial} (as presented in Table \ref{tab:phenotypes_b}). The study chooses to use Fitzpatrick Skin Types \cite{fitzpatrick1988validity} for skin tones as it provides sufficient granularity, \textit{\{Type 1, Type 2, Type 3, Type 4, Type 5, Type 6\}}, rather than binary skin-tone groupings, \textit{\{lighter skin-tone, darker skin-tone\}}. For eye shape, \cite{yucer2022measuring} consider epicanthal folds and eyelid difference as a more distinctive attribute for racial bias \cite{lee2000anchor}. The nose is categorised into two, wide and narrow, by examining the nasal breadth \cite{zhuang2010facial}. Hair texture is down-sampled from the eight categories of \cite{de2007shape} into three main hair texture types: straight, wavy, curly, in addition to bald. Additionally, \cite{yucer2022measuring} retains hair colour, as it is related to skin tone \cite{rees2003genetics}, with hair colour categories: red, grey, black, blonde, brown (see Table \ref{tab:phenotypes_b}).  

Compared to the prevalence of race or skin tone categories, phenotype-based groupings have received less attention across the racial bias literature to date, as they involve both skilled attribute labelling for dataset construction and a significantly more complex evaluation strategy due to the significant number of phenotype categories, and phenotype combinations present. To these ends, within a phenotype-based grouping strategy the concept of race is not represented by the difference across a single facial phenotype but rather a combination of varying phenotypic differences that differentiate a given subject's facial characteristics from another. As such, subsequently investigating the impact of such differences on face recognition performance becomes both more complex and time-consuming despite the improved comprehensiveness and quantification options that such a phenotype-based approach offers to the evaluation. On the other hand, it is essential to note when used, the correlation of phenotypical categories with more traditional (i.e. historically problematic, see Sec. \ref{sec:3:1}) racial categories should be avoided in order to prevent the naturalisation (or popularisation) of such \textit{"headline style"} summation of racial bias evaluation results.  

In conclusion, this section presents an alternative methodology for addressing racial bias (group fairness) within face recognition tasks. Whilst the face naturally conveys identity-related biometric information, it also inherently reflects a significant genetic and geographic relationship with race but these secondary relationships with race are not the primary concern for face recognition tasks. Instead, the group fairness objective within face recognition tasks is to ultimately ensure that it equity of performance across all subjects, regardless of subject racial grouping or facial phenotype characteristics. To these ends, it is necessary to avoid the inherited problem of racial and skin tone category usage within face recognition datasets and processing pipelines (Sec \ref{sec:3:1} \& Sec \ref{sec:3:2}), and instead adopt a more general option that facilitates quantifiable performance measurement without any explicit reference to such problematic concepts. By contrast, the use of facial phenotypes offers a viable alternative that, whilst not fully independent of earlier racial categorisation, offers significantly more granular insight within the quantification of racial bias spanning both skin tone and numerous other facial characteristics.

%% file: tables/phenotypes.tex
\begin{table}[ht]
\vspace{-.3cm}
\rowcolors{1}{rowzebra}{white}
\begin{subtable}[t]{.49\linewidth}
\centering
\vspace{0pt}
\setlength\tabcolsep{3pt}
\begin{tabularx}{\textwidth}{@{}ll@{}}
\toprule
\rowcolor {white}
\textbf{Facial Coding}                     & \textbf{Description}\\
\midrule
Schema 1  \cite{schendel1995anthropometry} & Craniofacial Distances\\
Schema 2  \cite{farkas2005international}   & Craniofacial Areas\\
Schema 3  \cite{ramanathan2006modeling}    & Craniofacial Ratios\\
Schema 4  \cite{liu2003facial}             & Facial Symmetry\\
Schema 5  \cite{porcheron2017facial}       & Facial Regions Contrast\\
Schema 6  \cite{chardon1991skin}           & ITA-based Skin Tones\\
Schema 7  \cite{dex}                       & Age Prediction\\
Schema 8  \cite{dex}                       & Gender Prediction\\
Schema 9  \cite{appen_2022}                & Subjective Age \& Gender Annotation\\
Schema 10 \cite{dlib}                      & Pose and Resolution\\
\bottomrule
\end{tabularx}
\Description{}
\caption{Summary of facial coding scheme analysis for the DiF dataset \cite{merler2019diversity}.}
\label{tab:phenotypes_a}
\end{subtable}%
\hspace{\fill}%
\begin{subtable}[t]{.49\linewidth}
\centering
\vspace{0pt}
\setlength\tabcolsep{3pt}
\begin{tabularx}{\textwidth}{@{}ll@{}}
\toprule
\rowcolor {white}
\textbf{Phenotype Attribute} & \textbf{Categories}\\
\midrule
Skin Tone \cite{fitzpatrick1988validity} & Type 1 / 2 / 3 / 4 / 5 / 6\\
Eyelid Type \cite{lee2000anchor} & Monolid / Other\\
Nose Shape \cite{zhuang2010facial} & Wide / Narrow\\
Lip Shape & Full / Small\\
Hair Type \cite{de2007shape} & Straight / Wavy / Curly / Bald\\
Hair Colour \cite{schneider2019use} & Red / Blonde / Brown / Black / Grey\\
\bottomrule
\end{tabularx}
\Description{}
\caption{Facial phenotype attributes and their categorisation by \cite{yucer2022measuring}.}
\label{tab:phenotypes_b}
\end{subtable}
\vspace{-0.3cm}
\end{table}

%% file: tex/3-TowardsRacialGroupFairness/4-conclusions.tex
\noindent
Overall, within this section we explore the race, skin tone and facial phenotype grouping strategies with regard to the \textit{group fairness criteria} for racial bias within face recognition. We critically review current grouping strategies in face recognition datasets spanning race, skin tone and facial phenotypes. Whilst race remains a controversial concept that carries \textit{historical bias}, ambiguity, ill-definition and disparity, a plethora of research identifies the possible risks and related problems of racial subject categorisation as a primary means for bias quantification within face recognition systems (Sec. \ref{sec:3:1}). Alternatively, whilst skin colour has been utilised to both quantify and address racial bias, it remains only one trait of what is a comprehensive and multi-faceted race concept (Sec. \ref{sec:3:2}). A broader approach, using facial phenotype as race-related facial attributes, provides a more objective and granular evaluation strategy for racial bias within face recognition (Sec. \ref{sec:3:3}). However, whilst the overall aim is to achieve more accurate and fairer face recognition system performance across increasingly more diverse populations, we need to still ensure the race and related interpretations are not reduced to only facial phenotypes by ignoring the broader context of cultural, historical and social factors \cite{hanna2020towards}. Moreover, the assessment of any grouping strategy on facial imagery creates another area of concern attributable to the often uncontrolled and inconsistent imaging conditions of facial capture that themselves lead to erroneous racial grouping annotation. As we move forward, we must address such risks, together with broader ethical considerations, within the wider development of face recognition processing pipelines.

%% file: tex/4-RacialBiasWFaceRecognition.tex
\input{tex/4-RacialBiasWFaceRecognition/0-intro.tex}

\input{tex/4-RacialBiasWFaceRecognition/1-image-acquisition.tex}

\input{tex/4-RacialBiasWFaceRecognition/2-face-localisation.tex}
\input{tex/4-RacialBiasWFaceRecognition/3-face-representation.tex}
\input{tex/4-RacialBiasWFaceRecognition/4-verification-identification.tex}

%% file: tex/4-RacialBiasWFaceRecognition/0-intro.tex
\vspace{-.4cm}
\section{Racial Bias within Face Recognition}
\label{sec:4}

Contemporary automated facial recognition encompasses a pipeline of multiple stage processing; image acquisition (for both dataset collation and deployment), face localisation, face representation, face verification and identification (final decision-making) \cite{kortli2020face,ali2021classical}.
\\
\noindent
\textbf{Image Acquisition} covers image capture from a wide range of devices such as smartphone cameras, webcams, high-end DSLR cameras and CCTV-style video surveillance cameras varying imaging conditions that span image resolution and compression, facial occlusion, facial pose, illumination, subject use of make-up/glasses/jewellery and facial expression. Furthermore it includes all stages of initial image pre-processing and formulation such as the demosaicing conversion to per-pixel RGB colour (from the Bayer pattern of the camera CMOS/CCD device), automatic colour and contrast correction (including processes such as automatic exposure control, white balance, automatic focus, brightness correction), pixel quantisation to a given bit-depth (e.g. RGB 8-bit colour) and compression. For data set collation, acquisition is complemented by a data curation such that differing imagery is sampled to select a subset of representative images that are ideally diverse and challenging enough to capture the full range of faces and imaging conditions that a face recognition system may encounter in real-world (\textit{"in-the-wild"}) deployment. These are then used to form the train $D_{train}$ and test $D_{test}$ datasets for system training and evaluation (as defined in Sec. \ref{sec:2}).

\noindent
\textbf{Face Localisation} consists of two sequential steps to process real-world, in-the-wild images that are captured under uncontrolled conditions and may hence exhibit variation across one or more of the aforementioned imaging conditions (typically: face off centre, rotated and of varying scale relative to the camera). The first step, face detection aims to identify a set of facial landmark locations (e.g. eye, mouth and nose endpoints, face boundaries in width and height) whilst the subsequent step of face alignment aims to correct for positional, rotational and scale variations to obtain a canonical facial image representation. This facial alignment step facilitates the use of the spatial correlation of facial features across both varying subjects and dataset image samples within the subsequent stage of face representation.
\\
\noindent
\textbf{Face Representation} involves optimisation the mapping function $f^*$ that projects a given face image sample into a feature embedding space, where the feature embedding vectors are both representative and distinctive for each subject. In order to select the optimal mapping function, $f^*$, a training process is performed via a training dataset, $D_{train}$, with reference to the minimisation of a loss function $\mathcal{L}$ that incites the use of a distinctive facial feature mapping (as defined in Sec. \ref{sec:2}). Consequently, $f^*$ provides mapping for both the curated training dataset, $D_{train}$, and unseen images in both test dataset, $D_{test}$, and any subsequent deployment.
\\
\noindent
\textbf{Face Verification and Identification} encompass the two most common decision-making (i.e. \textit{"end goal"}) tasks in face recognition. Face verification refers to a one-to-one matching operation to determine whether two facial images belong to the same individual (known subject case), and identification refers to a one-to-many matching operation to conversely identify a given individual against a set of reference images (unknown subject case). The optimal selection of mapping function, $f^*$, via the training process on training dataset, $D_{train}$, directly impacts the effectiveness of the feature embedding vectors such that the presence of both improved representational distinctiveness between differing subjects and also the robust representation of identical subjects under varying imaging conditions hence leads to improved face verification and identification performance.

\begin{figure}[t]
\centering
\resizebox{\linewidth}{!}{
\includegraphics[]{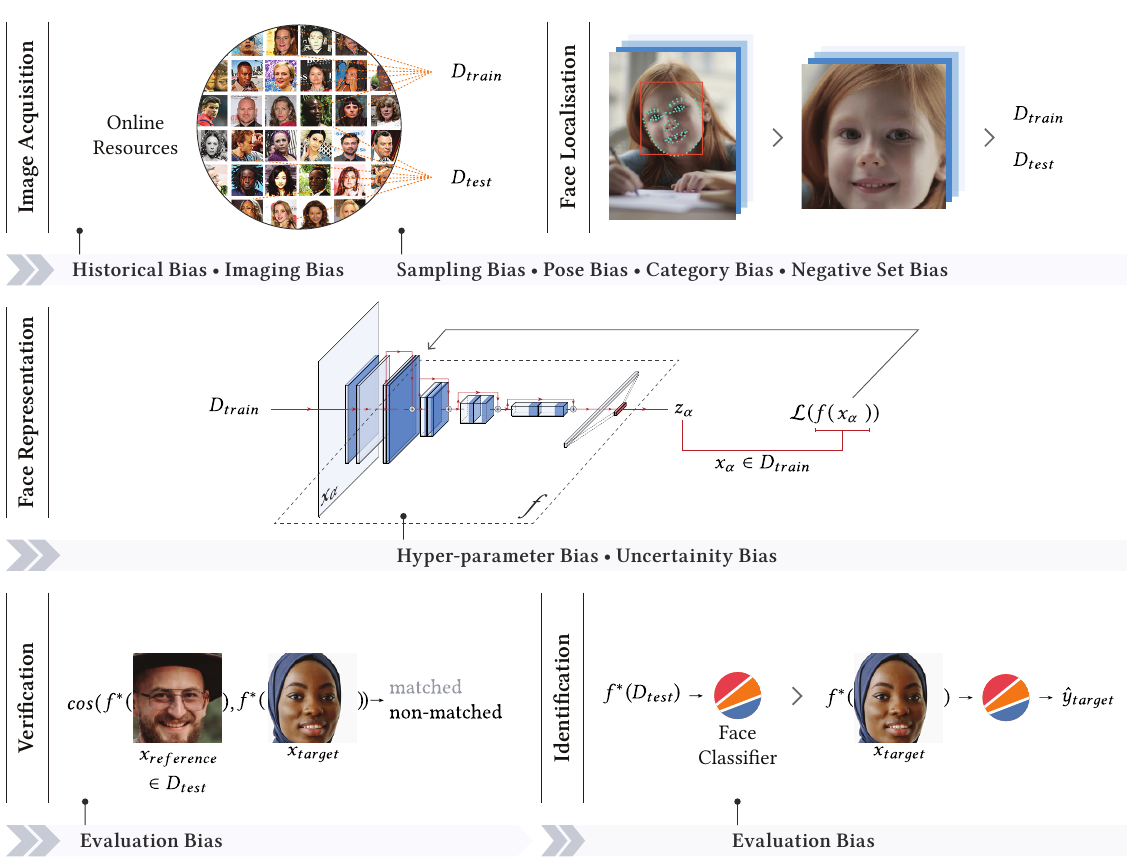}
}
\vspace{-0.4cm}
\caption{Overview of the face recognition processing pipeline and bias attribution.}
\vspace{-0.4cm}
\label{fig:overview}
\end{figure}

\noindent
With reference to the formal face recognition problem space definitions of Sec. \ref{sec:2}, this four stage conceptual face recognition processing pipeline is illustrated in Figure \ref{fig:overview} where we additionally highlight the potential sources of bias at each stage. These will be further explored, with reference to related work in the literature on racial bias within face recognition, in the remainder of this section.

%% file: tex/4-RacialBiasWFaceRecognition/1-image-acquisition.tex
\vspace{-.4cm}
\subsection{Image Acquisition}
\label{sec:4:1}

Image acquisition, spanning the imaging aspects of both initial dataset collation and final real-world deployment. We subdivide this stage into three categories, including \textit{facial imaging}, \textit{dataset curation}, and \textit{dataset bias mitigation}.

\vspace{-.3cm}
\subsubsection{\textbf{Facial Imaging}} Biometric data refers to distinctive physical characteristics of the human face, fingerprints, voice, iris, and body. Such biometrics have been used for identification systems (e.g. fingerprint matching) for several decades \cite{komarinski2005automated}. Commensurately, facial imagery has become a key part of modern biometric tasks due to the proliferation of imaging technologies, which significantly improve facial image quality, accessibility, and quantity. However, the increased prevalence of facial imagery does not necessarily result in improved biometric outcomes across all populations. In addition, collating facial images and annotating them with subject identity or racial category labels at scale have ignited complex discussions around policy and legality due to economic, privacy and ethical implications \cite{introna2010facial}. 

We have previously explored the \textit{historical bias} and potential risks associated with racial categorisation and the annotation of facial images (Sec. \ref{sec:3}). Building upon this, here we focus on the privacy risks and ethical concerns surrounding using facial images as a form of biometric data. Paying attention to such ethical and political considerations on the collation of biometric face imagery becomes particularly important when the presence of racial bias therein directly or indirectly impacts societal fairness. Accordingly, \cite{introna2010facial} presents a socio-political analysis of face recognition and highlights the distinct challenges and concerns associated with its development and evaluation. The study categorises such concerns into four sections: privacy, \textit{fairness}, freedom and autonomy, and security. Even though the intention of automatic face recognition is not problematic, in practice, it may enable morally unacceptable use cases of such technology. Examining the issue of subject consent, both within dataset collation and in an eventual use-case, is fundamental to that preserving privacy \cite{introna2010facial}. For example, government use of such technology for racial profiling and racially-targeted restriction in some jurisdictions has been widely reported \cite{gravett2020digital,wee2019china} and investigated \cite{daly2019algorithmic,van2020ethical}. In parallel to \cite{introna2010facial}, Prabhu \cite{birhane2021large} discusses the fundamentals of informed consent, privacy, or agency of the individual in large-scale datasets and shows the fallacy of the commonplace Creative Commons licensing model \cite{kim2007creative} as a consent-included green flag for large-scale dataset curation. They suggest the use of dataset audit cards as an approach to publishing the original research goals, curation procedures, known shortcomings and caveats alongside dataset dissemination \cite{birhane2021large}. Overall, it must be noted that any erosion of privacy, moral, ethical, or political values will most likely disproportionately impact minority groups, such as those defined along racial lines.

From a technical standpoint, the ISO/IEC 19794-5 \cite{iso19} standard and ICAO 9303 guidelines \cite{monnerat2007machine} propose both image-based (i.e. illumination, occlusion) and subject-based (i.e. pose, expression, accessories) image quality requirements to ensure facial image quality. Accordingly, facial images should be stored using lossy image compression standards such as JPEG \cite{JPEG} or JPEG2000 \cite{JPEG2000}; and observable in terms of gender, eye colour, hair colour, expression, properties (i.e. glasses), head pose (yaw, pitch, and roll), and facial landmark positions. However, commonplace \textit{"in-the-wild"} face datasets, that are readily used in face recognition system performance evaluation \cite{cao2018vggface2,guo2016ms}, do not conform to such requirements. Subsequently, Vangara  \cite{vangara2019characterizing} compares ICAO compliance between African and Caucasian groups in MORPH dataset \cite{ricanek2006morph} and found that slightly more than $48\%$ of the African-American images were rated as ICAO compliant, while slightly more than $57\%$ of Caucasian images were rated as ICAO compliant. The most prominent factor contributing to the variation in image quality between the groups is the difference in brightness; the distribution of which differs significantly between the African-American and Caucasian groups. The study argues that the lack of illumination correction with regard to skin tone during image acquisition could be the attributable reason as to why the African-American image group contains a larger number of poorly illuminated images. In parallel, \cite{cook2019demographic} points out the significant impact of skin reflectance across demographic subgroup performance with regard to face recognition and mentions that improved imaging acquisition systems (superior camera specification, lower motion blur, higher image contrast and stricter pose control) may significantly reduce or eliminate performance differences between such subgroups.

Furthermore, prior literature shows that non-ideal imaging conditions, including image blur, noise, distortion, occlusion and lossy compression, all have a considerable impact on the performance of face recognition \cite{karahan2016image,poyser2021impact,majumdar2021unravelling,yucer2020lossy}. Recently, \cite{majumdar2021unravelling} examined distorted test imagery impact on gender and skin tone categories (light vs. dark skin tone) using pre-trained DCNN-based face recognition models. As a result, the study \cite{majumdar2021unravelling} finds that the regions of interest used in the models shift towards less distinctive regions in the presence of distortions, resulting in unequal performance degradation among subgroups. Consequently, Yucer \cite{yucer2020lossy} finds that using lossy compressed facial test imagery decreases performance more significantly on specific phenotypes, including dark skin tone, wide nose, curly hair, and monolid eye when considered relative to a broader set of 21 phenotypic features. However, whilst the use of compressed imagery during training does make the resulting models more resilient and limits the performance degradation encountered, lower performance amongst these specific racially-aligned subgroups remains. Additionally, Yucer \cite{yucer2020lossy} find that removing chroma subsampling, which is itself a key lossy component of contemporary image compression schemes \textit{FMR}, improves face recognition performance for the specific phenotype categories which are otherwise more adversely affected by the use of lossy compression.

Consequently, we refer to these performance disparity effects within face recognition caused by variable imaging conditions as \textit{imaging bias} as illustrated in Figure \ref{fig:overview}. The limited literature on \textit{imaging bias} within face recognition to date makes it harder to identify the presence of such bias and align it to common underlying factors and conditions. On the other hand, state-of-the-art techniques for robust face recognition such as \cite{knoche2022octuplet} may help to mitigate such \textit{imaging bias} effects, via the use of a rich set of input variations aligned to phenotypic characteristics,  such as skin colour or other common facial phenotype variations \cite{roh2021sample}.

\vspace{-.3cm}
\subsubsection{\textbf{Dataset Curation}}
The following stage of image acquisition pertains to sampling the captured and processed facial images in order to create representative datasets for face recognition evaluation. Nevertheless, such a sampling process is often affected by sampling bias (also similar to selection, representation, or population bias) \cite{cochran1977sampling}, which significantly impacts racial bias in face recognition. \textit{Sampling bias}, referring to non-random selection over a population leading to a set of samples that do not fairly represent that population statistically, commonly occurs when facial images are curated from public online image resources, where the available population image distribution may not be representative of the actual societal population that the face recognition system will encounter in deployment. This is attributable to the fact that technology access is not globally or socio-economically homogeneous resulting in a skewed online image presence for a subset of the populous. Secondly, the most common approach for face recognition dataset collation is via targeted per-subject search for named individuals (commonly celebrities from the FreeBase listing) using public online image resources \cite{guo2016ms} (see Table \ref{tab:frdatasets}, which then results in a dataset of millions of subjects who have/had public attention. 

Even more concerning is that the subsampling decision from the FreeBase celebrity list is most often based on ranking all the subjects by their frequency of occurrence in the media, meaning that celebrities with greater global media coverage are more likely to be included in the dataset. This results in a biased convergence to a specific celebrity group, which is dominated by Western, European and American subjects. Moreover, this impact of sampling bias can be subsequently amplified during the later stage of feature representation learning due to an increased imbalance of phenotypic features which are themselves aligned to the dominant racial or demographic groupings present from the original dataset curation \cite{hellstrom2020bias}. For instance, a DCNN-based face recognition model utilising certain features, such as hair colour, to identify face subjects results in a bias towards a particular hairstyle or hair colour, causing less accurate performance on subjects with different hairstyles, hair colours, or accessories. 

Consequently, contemporary face recognition datasets are largely curated to provide large-scale coverage of differing face subjects images under a rich variation of \textit{"in the wild"} imaging conditions, with little consideration of the racially differentiating phenotypes of the underlying subject population. The two most widely used training datasets for face recognition - MS-Celeb-1M \cite{guo2016ms} and VGGFace2 \cite{cao2018vggface2} - contain 10 million and 3.3 million face images respectively, and are curated from the FreeBase celebrity list as shown in Table \ref{tab:frdatasets}. Similarly, the most common benchmark test sets for face recognition - LFW (Labeled Faces in the Wild) \cite{LFWTech}, CASIA-WebFace \cite{yi2014learning}, and MegaFace \cite{kemelmacher2016megaface} - are curated using online news (Yahoo), FreeBase celebrity and public online photo sharing resources (Flickr), respectively. 
Despite efforts to overcome sampling bias within face recognition datasets, such as the release of new datasets like the CASIA-Face-Africa \cite{muhammad2021casia}, a large-scale African face image database or the BUPT-Balanced dataset \cite{wang2020mitigating}, a large-scale racially balanced training set, the most prominent face recognition datasets used for face recognition evaluation still suffer from sampling bias with regard racial phenotypical population coverage.

\vspace{-.3cm}
\subsubsection{\textbf{Dataset Bias Mitigation}}

The prevailing assumption in machine learning is that the training set $D_{train}$ and test set $D_{test}$ are identically and independently distributed. However, this assumption is not valid for face recognition, and hence results in an out-of-distribution (OOD) problem. Accordingly, \cite{torralba2011unbiased} relates the OOD problem to \textit{"dataset bias"}. Although face recognition datasets should be representative of the real-world population in order to enable real-world face recognition model deployment, current state-of-the-art face recognition approaches remain closed-set methods, reflecting the world in a significantly biased way \cite{torralba2011unbiased}. Subsequently, \cite{torralba2011unbiased} groups dataset bias into four different types of bias:- 1) \textit{selection bias} is similar to our aforementioned concept of \textit{sampling bias}; 2) \textit{capture bias} occurs when the dataset imagery contains targets (faces) that have minimal spatial and illumination variation and can be related with \textit{pose bias} within face recognition context, as there is still poor pose variance (i.e. $\pm30^{\circ}$ horizontal, $\pm15^{\circ}$ vertical) within facial datasets; 3) \textit{category or label bias} poses the ill-definition or mislabelling of subject identities and racial categories; 4) \textit{negative set bias} defines bias against target appearances that are not represented in the data set (i.e. 
\textit{"the rest of the world"} appearance) leading to recognition models that are overconfident and misrepresent performance by considering only a skewed subset of possible real-world data samples (i.e. the test dataset, $D_{test}$).

In order to mitigate dataset bias, many studies \cite{kamiran2010classification,sadhukhan2019learning} propose novel sampling methods by either down-sampling or upsampling (i.e. augmenting) the datasets in the early stage of the face recognition processing pipeline. With the latest advancements in Generative Adversarial Networks (GAN) \cite{karras2020analyzing,xu2018fairgan,tan2020improving}, high-quality face image generation has become available as a potential tool to overcome the adverse effects of dataset distribution bias on subsequent real-world generalisation performance. For example, \cite{kortylewski2019analyzing} addresses dataset bias within face recognition via the use of synthetic dataset augmentation. The study shows that deeper DCNN architectures generalise better to unseen facial poses, then trained using synthetically augmented datasets, and hence the impact of dataset bias can be reduced by $75\%$. Another work \cite{yucer2020exploring} aims to automatically construct a synthesised dataset by transforming facial images across varying racial domains while preserving identity-related features, such that racially dependent features subsequently become irrelevant within the determination of subject identity. Similar to \cite{yucer2020exploring}, \cite{ge2020fgan} transforms the facial images of one racial category to corresponding images of other racial categories in order to facilitate a more balanced racial category distribution via data augmentation. Moreover, \cite{mroueh2021fair} proposes a new data augmentation strategy that imposes the fairness constraint to improve the generalisability of fair classifiers. In particular, they highlight that fairness can be achieved by augmenting interpolated samples between the racial groups during training. However, such generative models themselves produce samples from the underlying training set distribution upon which they are trained, meaning that they can also be impacted by dataset bias. Accordingly, \cite{tan2020improving} conduct an empirical study on the fairness of state-of-the-art pre-trained face synthesis GAN models. They show that a strong correlation between the imbalance in the original GAN training data and that of the resultant distribution of the GAN output images meaning that any dataset bias present is only amplified in cases where GAN are used as a potential data augmentation strategy for face recognition. 

Overall, this section outlines various sources of bias that can affect the accuracy and fairness of face recognition systems, such as \textit{imagery bias, sampling bias, pose (capture) bias, category and label bias, and negative set bias} as illustrated in Figure \ref{fig:overview} against the corresponding stages of the face recognition pipeline.

%% file: tex/4-RacialBiasWFaceRecognition/2-face-localisation.tex
\vspace{-.4cm}
\subsection{Face Localisation}
\label{sec:4:2}

The face localisation stage of the face recognition pipeline consists of face detection and alignment, thereby enabling the spatially correlated facial features for the subsequent stage of face representation. Prior work has primarily focused on hand-crafted facial feature extraction and classification for face detection. In a notable milestone, Viola and Jones proposed a real-time cascade of simple Haar-like feature classifiers at locally learned image locations \cite{viola2004robust}. Recently, face detection methods have shifted towards DCNN-based architectures and are categorised into five sub-genres by \cite{minaee2021going}: Cascade-CNN-based, R-CNN and Faster-RCNN-based, Single Shot Detection, Feature Pyramid Network-based, and other variants. Subsequently, the two most prominent face detectors, Cascade-CNN-based MTCNN \cite{zhang2016joint} and Feature Pyramid Network-based RetinaFace \cite{deng2020retinaface}, and the face detection benchmark dataset, Wider Face \cite{yang2016wider}, have become widely adopted for face recognition.

The MTCNN face detector is based on a cascading multi-tasking structure \cite{zhang2016joint} with three-stage lightweight DCNN where the Proposal Network (P-Net) generates a set of face regions, or "proposals", at different scales, the Refinement Network (R-Net) subsequently refines such regions to better localise the faces and finally the Output Network (O-Net) performs fine-grained face feature extraction and classification. Subsequently, \cite{deng2020retinaface} proposes another multi-level face localisation approach, RetinaFace, encompassing a single-shot detection network, a multi-task branch network that predicts both facial landmarks and attributes, and a bounding box regression network refines the position and size of the detected faces from the facial landmarks and attributes. Both approaches achieve outstanding performance on several benchmarks, including Wider Face \cite{yang2016wider}, which comprises 32,203 images and 393,703 bounding boxes under varying imaging conditions.


Despite the widespread usage of face detectors within the face recognition processing pipeline, only a few studies have investigated racial bias within face detection. Menezes \cite{menezes2021bias}, analysis the performances of five state-of-the-art face detectors; DSFD \cite{li2019dsfd}, Pyramid Box \cite{tang2018pyramidbox}, LFD \cite{he2019lffd}, RetinaFace \cite{deng2020retinaface}, MTCNN \cite{zhang2016joint} on demographic attributes including age, skin tone, gender. The study randomly samples the Casual Conversation Video Dataset \cite{hazirbas2021towards} and obtains 550.000 frames for training. The Casual Conversation Video Dataset adapts the Fitzpatrick scale and contains an imbalanced skin tone category distribution with the percentages of Skin Type 1: 4.0\%, Type 2: 28.3\%, Type 3: 22.9\%, Type 4: 8.4\%, Type 5: 15.8\%, Type 6: 20.7\%. Although Type 1 skin tone has the lowest representation in the training data, LFD, DSFD, and it was found that empirically RetinaFace detectors are more likely to fail to detect faces with skin type 4. Moreover, the study shows that the highest divergence of FNMR occurs within skin tone (being worse than age and gender groupings) and highlights that three out of five detectors evaluated have a higher likelihood of incorrect detection \textit{(FNMR)} for darker skin tones (Type 5 and 6).

Another study \cite{dooley2021robustness} investigates the robustness of three commercial online face detection capable systems: Amazon Rekognition, Microsoft Azure, and Google Cloud Platform and evaluates the impact of 15 types of natural noise corruption on the face detection performance of different demographic groups. Similarly to the case of face recognition, they conclude that corrupted data is more likely to cause face detection errors in specific demographic groups. For example, those with darker skin types, older adults, and those with masculine presentation all had higher errors ranging from 20-60\%. Subsequently, they compare the performance and robustness of non-commercial approaches (TinaFace \cite{zhu2020tinaface}, YOLO5Face \cite{qi2021yolo5face}, MogFace \cite{liu2022mogface}) with commercial ones \cite{dooley2022commercial}. They show that commercial approaches are always as biased or even more biased than non-commercial models, despite relatively larger development investment and supposed dedication to industry-level fairness commitments. More recently, \cite{mittal2023face} proposes the Fair Face Localisation with Attributes (F2LA) dataset with demographic annotations to detect disparate performance over such demographic groups. The study finds that confounding factors, including facial orientation, illumination, and resolution, can cause such disparate performance among demographic groups. Therefore it is important to analyse the performance of such detection models holistically and not draw conclusions solely based on demographic annotations. 

Despite ample evidence indicating the existence of racially disparate performance within face detection, there needs to be further investigation targeting racial bias exploration within face detection. Furthermore, similarly to the image acquisition stage of face recognition (Section \ref{sec:4:1}), the presence of imaging, sampling and dataset bias within these face detection benchmark datasets again translates through the subsequent stages of face recognition resulting in skewed overall face recognition pipeline performance.


%% file: tex/4-RacialBiasWFaceRecognition/3-face-representation.tex
\vspace{-.4cm}
\subsection{Face Representation}
\label{sec:4:3}

Facial feature representation has been a prominent area of computer vision research for many decades and several milestones have substantially improved the performance of face recognition today \cite{wang2021deep}. The first well-known method for estimating the probability of distribution over high-dimensional vector space of face images, Eigenface, was introduced in the early 1990s \cite{turk1991eigenfaces}. Following that, Gabor \cite{liu2002gabor} and LBP \cite{ahonen2006face} provide robust performance by using local filtering to obtain invariant facial features. However, they could not create handcrafted features that were distinctive and compact enough to fully scale to the diversity of large-scale benchmark datasets (and hence the global populous). Although numerous learning-based local descriptors have been developed to tackle various aspects of face recognition  \cite{cao2010face,lei2013learning}, higher similarity for intra-class samples and diversity for inter-class samples within face datasets remain challenging. Subsequently, the availability of large-scale dataset resources (2007+) and the proliferation of DCNN (2012+) have now enabled contemporary face recognition architectures to achieve outstanding verification and identification accuracy. Accordingly, this stage involves a mapping operation from face images to face representation vectors which can be performed by a DCNN-based backbone architecture and a loss function, as discussed in Sec.\ref{sec:2}.

\vspace{-0.3cm}
\subsubsection{\textbf{Backbone Architectures.}}  
DCNN are multi-layer processing blocks, including convolutional, pooling  and fully connected layers. As a central component of DCNN, the convolutional layers extract features from the output of the previous layer, starting from the face image input. Each layer $t$ consists of $K$ kernels with weights $W = W_1, W_2,..., W_K$ and added bias filters $B = b_1,...,b_K$. Subsequently, each layer applies an element-wise nonlinear transform (i.e. $\sigma \in \{RELU, tanh, Softmax, \dots\}$ functions) to generate multiple feature map representations and passes the result to the next layer ${x}^{t} = \sigma(W_k \cdot x^{t-1} + b_k)$. Moreover, at the end of each layer, a pooling function down-samples the feature maps by taking the maximum or average value of adjacent pixels (patch). Similarly, a fully connected layer applies a linear transformation to the input vector through a weights matrix. 
\\
\noindent
A majority of face recognition methods adopt state-of-the-art DCNN as their backbone architectures, such as the VGG-Net \cite{simonyan2014very}, the ResNet \cite{he2016deep}, and the Inception-ResNet \cite{schroff2015facenet}. VGG-Net \cite{simonyan2014very} uses a smaller fixed number of convolutional filters compared to the AlexNet \cite{krizhevsky2017imagenet} to decrease the total number of trained parameters. On the other hand, ResNet \cite{he2016deep} uses skip connections between two consequent layers to avoid the vanishing gradient problem (unstable training of deep networks due to ever decreasing gradients relative to the input). Furthermore, InceptionNet \cite{schroff2015facenet} consists of multiple kernels in one layer to grasp salient features at different levels, including global and distributed features.


\vspace{-0.4cm}
\subsubsection{\textbf{Baseline Loss Functions}}
Contemporary, face recognition literature primarily focuses on designing novel DCNN loss functions \cite{liu2017sphereface,wang2018cosface,deng2019arcface,cao2018vggface2} to enhance the distinctiveness and separability of features. Mostly, such loss functions \cite{liu2017sphereface,wang2018cosface,deng2019arcface} operate on the feature embedding vectors of the last fully connected layer of the selected backbone DCNN architecture \cite{he2016deep}. Previously, we discussed Softmax loss $\mathcal{L}_{softmax}$ (Eqn. \ref{eq:softmax}) which is based on maximising the posterior probability of the ground-truth subject class in order to separate features from different classes. However, a high number of subject identities, $n$, within training sets increases the size of the linear transformation matrix in the last layer $W \in R^{d \times n}$ leading to high complexity. Moreover, the learned feature embedding vectors of Softmax loss are not distinctive enough to address the open-set face recognition problem \cite{he2019softmax}. To address these problems, CosFace \cite{wang2018cosface} enforces a larger cosine margin $m$ between the features of different classes and suggests that both norm of the vectors contribute to the posterior probability. 
\vspace{-0.1cm}
\begin{align} \label{eq:cosface}
\mathcal{L}_{cosface} = - \frac{1}{N} \sum_{i=1}^{N}  \log \frac{e^{ \lVert z \rVert (\cos(\theta_{y_i,i})-m)}}{e^{\lVert z \rVert(\cos(\theta_{y_i,i})-m)} + \sum_{j \neq y_i}^{n}{e^{\lVert z \rVert\cos(\theta_{j,i}))}}}
\quad  where \quad \cos({\theta}_j,i)={W^T}_j z_i
\end{align} where N is the number of training samples, $x_i$ is the $ith$ feature vector corresponding to the ground-truth class of $y_i$, the $W_j$ is the weight matrix of the $jth$ class, and ${\theta}_j$ is the angle between $W_j$ and $z_i$. Additionally, the bias term is removed $b=0$, and the weights $W$ and embeddings $z$ are normalised using $L_2$ normalisation. 

An alternative loss function, ArcFace \cite{deng2019arcface} differs from CosFace \cite{wang2018cosface} based on its distinct margin $m$. ArcFace has a more accurate geodesic distance because it has a constant linear, angular margin $m$ penalty throughout the interval, while CosFace has a nonlinear angular margin. Similarly, it normalises the weights and embeddings and fixes the bias term to zero. The ArcFace loss function is formalised as follows:
\begin{align} \label{eq:arcface}
\mathcal{L}_{arcface} = - \frac{1}{N} \sum_{i=1}^{N} \log \frac{e^{\lVert z \rVert(\cos(\theta_{y_i,i}+m))}}{e^{\lVert z \rVert(\cos(\theta_{y_i,i}+m))} + \sum_{j \neq y_i}^{n}{e^{\lVert z \rVert(\cos(\theta_{j,i}))}}}
\end{align}
where all definitions are as per Eqn. \ref{eq:cosface}. Overall the key Softmax, CosFace \cite{wang2018cosface}, and ArcFace \cite{deng2019arcface} differences lie in their use of deep face representation, weight vectors and their margin penalty in the last layer. Consequently, the accuracy of the most popular LFW benchmark has increased from $\sim60\%$ (Eigenfaces , \cite{turk1991eigenfaces} (1991)) to above $\sim99\%$ (ArcFace \cite{deng2019arcface} (2019)) further encouraging the broader adaption of face recognition into real-world applications. 
\input{tables/rfw}

The central concept of statistical learning is based on the requirement to choose one generalisation over another in order to be able to classify instances non-arbitrarily beyond those in the training set \cite{mitchell1980need}. Moreover, \cite{mitchell1980need} defines unbiased generalisation as one which makes no prior assumptions about which classes of instances are most likely to occur and bases all its decisions solely on data observation. However, any face recognition system already has dataset bias, meaning that any type of generalisation or observation based on such datasets results in bias. On the other hand, \cite{hellstrom2020bias} identify two more different type of bias occurs in this face representation stage. The study, first, mentions DCNN \textit{hyper-parameter bias} due to the ubiquitous number of hyper-parameters which are spanning from the choices of number of hidden nodes and layers to type of activation functions made by the user \cite{bertrand2019hyper}. The strong influence of such chosen parameters on DCNN and their performance makes \textit{hyper-parameter bias} relevant to racial bias as such in the case of \textit{hyper-parameter bias}, certain models may perform better on datasets that are biased towards certain groups leading to potentially perpetuating racial bias. \textit{Hyper-parameter bias} can also be related with aggregation bias (causing selected parameters forming the mapping function is not optimal for specific groups) defined by \cite{suresh2019framework}. Another type of bias, denoted as \textit{uncertainty bias}, is based on the probability values that are often computed together with each produced DCNN architecture. The probability represents uncertainty, and typically has to be above a set threshold for face detection, verification or identification to be performed. For example, a DCNN-based face detection model reports detection predictions via probability values indicating detection confidence. However, this manual selection of the probability threshold can itself create a bias when the threshold is set too conservatively such that faces from underrepresented groups are be more likely to not be detected due to higher uncertainty in the model. 
\noindent
Up to this point, we have described the general processes within the face representation stage of a face recognition architecture (Fig. \ref{fig:overview}) and the various forms of bias that may exacerbate racial bias within them. Finally, we complete our discussion of facial representation by exploring current racial bias mitigation strategies and categorise them into three sub-genres:- mutual information mitigation (Sec. \ref{sec:4:3:3}), loss function based mitigation (Sec. \ref{sec:4:3:4}), and domain adaptation based mitigation (Sec. \ref{sec:4:3:5}).

\vspace{-.4cm}
\subsubsection{\textbf{Mutual Information Mitigation}}
\label{sec:4:3:3}
The high mutual information between facial identity and underlying racial features within face images generally transfer into the learned feature embedding of contemporary DCNN based techniques and hence results in an unsatisfied \textit{fairness through unawareness} criteria (i.e. the constraint of not retaining information related to $s$ when estimating $y$ as the the formalised problem statement of Sec. \ref{sec:2}). A myriad of studies \cite{end,debface,creager2019flexibly,learn_not_to,sensitivenets,Robinson2021,ragonesi2020learning,dhar2021pass} attempt to decrease this mutual information in order to debias the performance of face recognition approaches. For example, \cite{end} provides a general framework with a regularisation strategy such that a model trained on a dataset that is known to be bias \textit{a priori} can be trained in to avoid the selection of biased features therein. The information bottleneck in the model distills the biased features (such as texture, background) and correctly learns to focus on relevant features (such as shape, e.g. within biased MNIST \cite{end}). Moreover, \cite{creager2019flexibly} proposes a Flexibly Fair VAE (FFVAE) algorithm concerning demographic parity among multiple sensitive attributes. FFVAE learns the encoder distribution from input and sensitive attributes and disentangles prior structure in latent space by enforcing low mutual information. On the other hand, adversarial-debiasing approaches become applicable in disentangling race-related information on faces within generative generator-discriminator models such as GAN \cite{dhar2021pass,debface}. For example, the Protected Attribute Suppression System (PASS) \cite{dhar2021pass} discourages the generator from encoding information related with sensitive attributes via discriminator. Furthermore, \cite{learn_not_to} uses a feature mapping network to unlearn biased sensitive attributes in order to disentangle the mutual information between identity and sensitive characteristics. Similarly, \cite{sensitivenets} suppress the presence of sensitive information to enforce the learning of privacy-preserving embeddings (for any sensitive feature we want to protect) and hence equality across such sensitive attributes in any subsequent decision-making algorithms based on these embeddings. Their results show that it is possible to reduce the performance of gender and ethnicity detection by 60-80\% on a given facial image embedding, while face verification performance over the same embedding is only impacted by 5\% .

Other recent works on mitigating racial bias introduce a knowledge distillation module for face recognition \cite{jung21,distill_debias, rectifying}. Accordingly, \cite{distill_debias} observes that the face recognition networks attend to different spatial regions in faces according to the category of an attribute label (e.g. light skin vs. dark skin tone). Firstly, in order to eliminate differences in the representations, they propose a teacher-student network that enforces to student network to generate teacher-like representations. Whilst the teacher network is trained on light skin tone images, the student network is trained on dark skin tone images. However, forcing student networks to attend only teacher networks spatial regions does not give fairer results than attending both spatial regions. As a result, they achieve less biased results in face verification and perform better than state-of-the-art adversarial debiasing approaches. Another study, \cite{rectifying} applies knowledge distillation from teacher to student to avoid dataset bias which is identified as an imbalance distribution between either class labels or between easy and hard dataset samples. The imbalance between samples decreases the uniformity of the data, which subsequently makes the data distribution far from uniform. As image datasets are usually collected ad-hoc without any inherent uniformity consideration, they propose two different sampling methods, extrinsic sampling (before training) and intrinsic sampling (during training), to ensure the success of knowledge distillation. On the other hand, some experiments empirically demonstrate that the use of race related facial feature increases overall face classification performance and improves extracted feature discriminability \cite{ryu2017inclusivefacenet}. 

\vspace{-.4cm}
\subsubsection{\textbf{Loss Function Based Mitigation}} 
\label{sec:4:3:4}
Another area of study \cite{yang2021ramface,wang2021meta,qin2020asymmetric} focuses on setting adaptive margins to tackle racial bias. For previous face recognition baselines \cite{wang2018cosface,deng2019arcface}, the margin between classes was set at a fixed value to maximise accuracy. However, the training distributions of demographic groups and their feature embedding vectors inherently differ from each other meaning that a global margin is essentially a best fit to the largest demographic group in the training dataset. While such a constant global margin may result better performance across one demographic, that same margin may conversely cause inferior performance for another.

Recently, \cite{yang2021ramface} proposes \textit{Race Adaptive Margin (RAM) Loss} using a new compact margin instead of using an ArcFace-style fixed margin, $m$ (Eqn. \ref{eq:arcface}), approach. Consequently, they define intra-subject compactness $\mu^r_{intra}$ for each racial group, $\{African, Asian, Indian, Caucasian\}$, in the RFW dataset in order to assign the margin to be an identity-related parameter.
As such, the final RAM Loss (denoted ramface loss, \cite{yang2021ramface}) is; 

\begin{align}
\label{eq:ramface}
\begin{array}{l}
\mathcal{L}_{ramface} = - \frac{1}{N} \sum_{i=1}^{N} \log \frac{e^{\lVert z \rVert(\cos(\theta_{y_i,i}+m_r))}}{e^{\lVert z \rVert(\cos(\theta_{y_i,i}+m_r))} + \sum_{j \neq y_i}^{n}{e^{\lVert z \rVert(\cos(\theta_{j,i}))}}} \\ where \quad  m_r = \beta \times Z_r \times \hat{\mu}^r_{intra}, \quad
\mu^r_{intra} =\frac{1}{B_r} \sum^{B_r}_{y=1}\frac{1}{M_{y_j}} \sum^{M_{y_j}}_{i=1} \cos\theta^r_{z_i^{y_j},c_{y_j}}.
\end{array}
\end{align}
where $B_r$ is the number of subject identities in the race group, $M_{y_j}$ is the number of the samples with subject class $y_j$, $Z_r$ is the race classification accuracy as the weight indicator in the adaptive margin loss, and $\beta$ is the scaling parameter to constrain the upper bound of $m_r$. As per ArcFace loss, Eqn. \ref{eq:arcface}, $z_i$ is the feature representation of image $x_i$.
Consequently, they benefit from racially-aware supervision to increase the distinctiveness of the learned feature representations and simultaneously decrease the potential for racial bias within that same representation. RamFace Loss achieves both high accuracy on face verification and appears to successfully mitigate racial bias (see Tab. \ref{tab:rfw}).

Another study, \cite{qin2020asymmetric} proposes an \textit{Asymmetric Rejection Loss}, which aims to reduce the racial bias within trained face recognition models by taking advantage of unlabelled images of under-represented groups. The study utilise unlabelled images collected from online sources where the number of subject identities present is always much greater than the average images per subject. Subsequently, they consider each unlabelled image as a separate class and design an asymmetric learning procedure for those labelled and unlabelled images.
Their proposed \textit{Asymmetric Rejection Loss} (denoted \textit{arl}) is defined as:
\vspace{-0.1cm}
\begin{align}
\mathcal{L}_{arl}= \mathcal{L}_L +\lambda_U \times \mathcal{L}_U + \lambda_C \times \mathcal{L}_C
\quad
where \quad
\mathcal{L}_C = \frac{\sum_{i,j}cos(z_i,z_j)^2}{N_t}, 0<cos(z_i,z_j)<t
\end{align}
where $t$ is the upper bound of the penalty interval, and $N_t$ is the number of feature representation vectors pairs whose cosine similarity lies within the interval $(0,t)$. $L_L$ and $L_U$ are similar to ArcFace loss equation \ref{eq:arcface} operating on labelled and unlabelled images respectively. Simultaneously, $\lambda_U$ and $\lambda_C$ are two loss weights. \textit{Asymmetric Rejection Loss} achieves improved performance on under-represented demographic groups whilst performance on well-represented groups remains unaffected when compared to other state-of-the-art approaches (Tab. \ref{tab:rfw}).

\vspace{-.3cm}
\subsubsection{\textbf{Domain Adaptation Based Mitigation}} 
\label{sec:4:3:5}
Following from the discussion of Section \ref{sec:4:1} on the out-of-distribution problem, domain adaptation techniques have recently been introduced as a method to address racial bias issues \cite{wang2019racial, yin2019feature, guo2020learning,faraki2021cross, Nam_2021_CVPR}. These techniques use multiple labelled source domains with different distributions to improve generalisation to new target datasets. One of the first examples of domain adaptation for racial bias, \cite{wang2019racial} prove the domain gap between racial groupings and propose a deep information maximisation adaptation network (IMAN) architecture to address this. Subsequently, \cite{guo2020learning} propose a novel face recognition methodology via the use of meta-learning named Meta Face Recognition (MFR). The meta-optimisation objective of MFR first synthesises the source/target domain. Subsequently, it forces the model to learn effective representations of both synthesised source and target domains. In another example in face recognition, \cite{faraki2021cross} introduces Cross-Domain Triplet (CDT) loss based on the triplet loss \cite{schroff2015facenet} and uses similarity metrics from one domain to learn compact feature clusters of identities by incorporating them into another domain. Relative performance for both CDT and MFR on the RFW dataset are shown in Table \ref{tab:rfw}.
\\
This section presents a brief overview of face representation learning, including the potential sources of biases and mitigation studies within this stage of the face recognition processing pipeline (Fig. \ref{fig:overview}). In support of this review of prior work on racial bias mitigation a summary table of related work is provided to compare overall relative performance on the RFW dataset \cite{wang2020mitigating} (Table \ref{tab:rfw}). 





  



%% file: tables/rfw.tex
\begin{table}[htbp]
\rowcolors{1}{rowzebra}{white}

\begin{tabular}{lllcccccc}
\toprule
\rowcolor {white}
\textbf{Methods} & \textbf{Backbone} & \textbf{Dataset} & \textbf{African} & \textbf{Asian} & \textbf{Caucasian} & \textbf{Indian} & \textbf{Avg} & \textbf{STD} \\ 
\midrule\hiderowcolors
\multicolumn{9}{c}{\textbf{\textbf{Imbalanced Training Sets}}}\\
\hline
\noalign{\global\rownum=1}
\showrowcolors
ArcFace \cite{deng2019arcface}                  & ResNet-34    & MegaFace      & 85.13 & 86.27 & 94.78 & 90.48 & 89.17 &4.39 \\
IMAN-A \cite{wang2019racial}                    & ResNet-34    & MegaFace      & 91.42 & 91.15 & 94.78 & 94.15 & 92.88 &1.86 \\
ArcFace \cite{deng2019arcface}                  & ResNet-34    & VGGFace2      & 87.30 & 85.47 & 93.50 & 87.55 & 88.46 &3.49 \\
ARL+C \cite{qin2020asymmetric}                  & ResNet-34    & VGGFace2      & 88.57 & 87.65 & 93.48 & 89.35 & 89.76 & 2.57\\
ArcFace \cite{deng2019arcface}                  & ResNet-50    & BUPT-Global   & 96.28 & 96.03 & 98.22 & 96.77 & 96.83 & 0.98\\
MV-Softmax \cite{wang2020mis}                   & ResNet-50    & BUPT-Global   & 95.83 & 95.66 & 99.33 & 95.83 & 96.66 & 1.78\\
DebFace-ID \cite{debface}                       & ResNet-50    & BUPT-Global   & 93.67 & 94.33 & 95.95 & 94.78 & 94.68 &0.96 \\
CurricularFace \cite{huang2020curricularface}   & ResNet-50    & BUPT-Global   & 94.93 & 95.18 & 97.75 & 96.07 & 95.98 &1.28 \\
RamFace \cite{yang2021ramface}                  & ResNet-50    & BUPT-Global   & 96.73 & 96.17 & 98.28 & 96.77 & 96.99 &0.90 \\
ArcFace \cite{deng2019arcface}                  & ResNet-101   & VGGFace2      & 89.45 & 87.61 & 94.71 & 91.21 & 90.75 &3.02 \\
VGGF2 Races \cite{yucer2020exploring}           & ResNet-101   & VGGFace2      & 90.10 & 87.73 & 93.72 & 90.50 & 90.51 & 2.46\\
ArcFace \cite{deng2019arcface}                  & ResNet-101   & BUPT-Global   & 96.77 & 96.52 & 98.55 & 97.48 & 97.33 & 0.91\\
CurricularFace \cite{huang2020curricularface}   & ResNet-101   & BUPT-Global   & 96.30 & 95.98 & 97.83 & 96.70 & 96.70 &0.81 \\
RamFace \cite{yang2021ramface}                  & ResNet-101   & BUPT-Global   & 97.40 & 96.93 & 98.65 & 97.57 & 97.64 & 0.73\\ 
\midrule\hiderowcolors
\multicolumn{9}{c}{\textbf{\textbf{Balanced Training Sets}}}\\
\hline
\noalign{\global\rownum=1}
\showrowcolors
Softmax                                         & ResNet-34    & BUPT-Balanced & 91.42 & 91.23 & 94.18 & 92.82 & 92.41 & 1.38\\
CosFace \cite{wang2018cosface}                  & ResNet-34    & BUPT-Balanced & 92.98 & 92.98 & 95.12 & 93.93 & 93.75 & 1.02\\
ArcFace \cite{deng2019arcface}                  & ResNet-34    & BUPT-Balanced & 93.98 & 93.72 & 96.18 & 94.67 & 94.64 & 1.10\\
RL-RBN \cite{wang2020mitigating}                & ResNet-34    & BUPT-Balanced & 95.00 & 94.82 & 96.27 & 94.68 & 95.19 &0.73 \\
RamFace \cite{yang2021ramface}                  & ResNet-34    & BUPT-Balanced & 95.28 & 94.83 & 97.15 & 96.08 & 95.84 & 1.02\\
GAC-ArcFace \cite{gong2021mitigating}           & ResNet-34    & BUPT-Balanced & 94.12 & 94.10 & 96.02 & 94.22 & 94.62 & 0.94\\
Fairness FR \cite{xu2021consistent}             & ResNet-34    & BUPT-Balanced & 95.95 & 95.17 & 96.78 & 96.38 & 96.07 & 0.69\\
ArcFace \cite{deng2019arcface}                  & ResNet-50    & BUPT-Balanced & 96.00 &	95.45 &	97.57&	96.42&	96.36&	0.90\\
CurricularFace \cite{huang2020curricularface}   & ResNet-50    & BUPT-Balanced & 94.90 & 94.23 & 96.38 & 95.50 & 95.25 &0.91 \\
RamFace \cite{yang2021ramface}                  & ResNet-50    & BUPT-Balanced & 96.25 & 95.50 & 97.40 & 96.58 & 96.43 & 0.79\\
GAC \cite{gong2021mitigating}                   & ResNet-50    & BUPT-Balanced & 94.65 & 94.93 & 96.23 & 95.12 & 95.23 & 0.69\\
Sensitive Loss \cite{serna2022sensitive}        & ResNet-50    & BUPT-Balanced & 95.82 & 96.50 & 97.23 & 96.95 & 96.63 & 0.62\\
Fairness FR \cite{xu2021consistent}             & ResNet-50    & BUPT-Balanced & 96.47 & 95.75 & 97.08 & 96.77 & 96.52 & 0.57 \\ \bottomrule
\end{tabular}

\caption{Performance of state-of-the-art face verification methods on the RFW dataset \cite{wang2019racial}, with comparison based on sample standard deviation.}
\Description{}
\label{tab:rfw}
\vspace{-0.9cm}
\end{table}

%% file: tex/4-RacialBiasWFaceRecognition/4-verification-identification.tex
\vspace{-.2cm}
\subsection{Face Verification and Identification}
\label{sec:4:4}

\noindent
The overarching concept of \textit{face recognition}, whereby an identity confirmation decision is made for a given subject based on facial images, can itself be subdivided into two discrete problems:- Face Verification (i.e. one-to-one facial comparison, Sec. \ref{sec:4:4:1}) and Face Identification (i.e. one-to-many facial comparison, Sec. \ref{sec:4:4:2}).

\vspace{-.2cm}
\noindent
\subsubsection{\textbf{Face Verification}}
\label{sec:4:4:1}
Face Verification refers to one-to-one facial comparison to verify the identity of a subject by comparing a hitherto unseen facial image against another \textit{a priori} image of the same or different subject. This is commonly used in access control systems for both physical locations (e.g. government sites, border control) and digital assets (e.g. smart phones, digital banking applications) hence representing the most common occurrence of a \textit{face recognition} technology encountered by the general public in contemporary society. Typically, face verification performance is measured in terms of accuracy (see Eqn. \ref{eq:acc}) and matching rates (see Eqn. \ref{eq:mr}) over pairs of identical/non-identical subject images in order to evaluate the number of correct identities matches over all the set of all paired images presented. In order to confirm a match, the feature embedding vector $z_{target}$ from a presented unseen subject image instance $x_{target}$, and those of a subject image $x_{reference}$ held on record \textit{a priori}, $z_{reference}$, are compared using a distance or similarity score across the learnt feature embedding space (e.g. cosine similarity). Subsequently, an a\textit{priori} $threshold$ is used to make a decision on the similarity of $z_{target} \approx z_{reference}$ such that a verified identity can be confirmed or not. Several studies demonstrate significant performance on face verification on public benchmark datasets \cite{maze2018iarpa, LFWTech} where the racial diversity within these datasets is often limited, biased and overlooked \cite{zhou2015naive}. Accordingly, the Labelled Faces in-the-wild Dataset (LFW) \cite{LFWTech} contains 13233 images of 1680 subjects, and 6000 specific pairs of images of subjects to measure 1:1 verification performance have become widely adopted. Subsequently, prior work \cite{deng2019arcface,wang2018cosface} has reached over 99.5 \% verification accuracy on LFW.

\vspace{-.4cm}
\noindent
\subsubsection{\textbf{Face Identification}}
\label{sec:4:4:2}
Face identification refers to a one-to-many facial comparison to identify an unknown facial query image by matching it to against a set of known facial images. Prototypically, law enforcement agencies use it to identify suspects in criminal investigations, track individuals in public spaces and search for missing persons. The process involves comparing an obtained query face image $x_{target}$ with a large database of reference images $X_{enrolment}$. Unlike face verification, which is used to verify the identity of a known individual, face identification is used to identify unknown individuals by matching their facial image to a reference image within the enrolment set for which the identity is known \textit{ a priori}. Face identification tasks can be sub-categorised as either closed-set, when the target is always in the enrolment set ($x_{target} \in X_{enrolment}$), or open-set, when the target may or may not be in the enrolment set ($x_{target} \in  X_{enrolment}$ or $x_{target} \not\in  X_{enrolment}$). Whilst the closed-set face identification task is limited to identifying only the subjects in its enrolment set, the more challenging task of open-set face identification is able to determine unknown faces that are not in the enrolment set. In order to perform a closed-set face identification task, a multi-class classifier is used to identify the target image $x_{target}$ via the use of feature embedding vector $z_{target}$ over $Z_{enrolment}$. Furthermore, for an open-set face identification task an additional threshold becomes necessary in order to ascertain an unknown target that is not present in the enrolment set. As for face identification, \cite{kemelmacher2016megaface} provides two large-scale face identification benchmark datasets under various imaging conditions.

Furthermore, \cite{suresh2019framework} defines \textit{evaluation bias} when the benchmark dataset used to post-training performance evaluation is not accurately representative of the target population (in deployment). The most common face recognition benchmark datasets \cite{LFWTech,grother2017ijb} illustrate examples of such evaluation bias, encouraging the development of models that only perform well on the specific racial groupings as the per distribution of the dataset (see Sec. \ref{sec:4:1}). Evaluation bias is also related to the decisions made at this stage of the face recognition pipeline, including pairing selection, threshold optimisation, distance and normalisation functions. For example, the selected threshold can vary across datasets, and final model performance is often susceptible to the changes in these thresholds  \cite{liu2022oneface}. Studies have found that a single fixed threshold often causes higher variance across demographic groups than an adaptive threshold per-group threshold \cite{liu2022oneface}. Another example, \cite{crosswhite2018template}, investigates template-based face verification and identification and the effects of template size, negative set construction and classifier fusion on performance. They find that performance is highly dependent on the number of images available in a template. Subsequently, \cite{krishnapriya2020issues} compares the accuracy for African-Americans and Caucasians, in a scenario in which a fixed decision threshold is used for all subjects only to find that African-Americans have a higher FMR and Caucasians have a higher FNMR.

Accordingly, many studies provide verification protocols and a new set of pairings based on racial groupings to address racial bias. For example, the following study \cite{wang2019racial} released the RFW dataset with a similar protocol to LFW \cite{LFWTech} with the same number (6000) of pairings for each of the four racial groups {African, Asian, Caucasian, Indian} with separate thresholds. Another study \cite{yucer2022measuring} annotates RFW for face verification and VGGFace2 for face identification with facial phenotype attributes to measure racial bias. Moreover, \cite{dhar2020adversarial} proposes the Adversarial Gender De-biasing algorithm (AGENDA) to train a shallow network that removes the gender information of the embeddings extracted from a pre-trained network. The authors of \cite{dhar2021pass} extend this work with PASS to deal with any sensitive attribute and proposed a novel discriminator training strategy. Subsequently, \cite{Terhorst20} (2020a) proposed the Fair Template Comparison (FTC) method, which replaces the computation of the cosine similarity score by an additional shallow neural network trained using cross-entropy loss, with a fairness penalisation and L2 penalty term to prevent over-fitting. While this method reduces model bias, it results in an overall decrease in accuracy and requires training and tuning of the shallow neural network. Another work, \cite{robinson2020face}, proposes a group-specific threshold (GST) in which the sensitive attributes themselves define its calibration sets. Another study, \cite{terhorst2020post} proposes the Fair Score Normalisation (FSN) method, which is essentially GST with unsupervised clusters. FSN normalises the scores by requiring the model FMRs across unsupervised clusters to be the same predefined global FMR. Salvador, \cite{salvador2022faircal} proposes a Fairness Calibration (FairCal) method that applies the K-means algorithm to the image feature representation vectors $Z$ and makes partitions of the embedding space into $K$ clusters. For each set, it calculates separate calibration map scores to cluster-conditional probabilities of the set. If the pair of images belong to the same subject cluster, the algorithm uses the score; if not, it uses the weighted average of the calibrated scores in each cluster of corresponding image features. Consequently, they achieve better overall accuracy, reducing the discrepancy in the FMRs while not requiring the use of the sensitive attribute.

Similar to face verification, open-set face identification requires a threshold to report a match or non-matched decision over test target imagery. Accordingly, \cite{krishnapriya2020issues}, highlight the importance of two types of errors in face identification false-non-matched identification and false-matched identification together with their dependency on a threshold that defines the minimum similarity required to report a match. Furthermore, \cite{yucer2022measuring} perform closed-set identification on VGGFace2 test set and show that performance difference between facial phenotypes is much smaller when compared to the face verification results.  However, the study is unable to have the same proportion for each attribute, and does not measure open-set face identification. Consequently, there is a need for the design and application of open-set tests for face identification using more diverse benchmark datasets and novel evaluation strategies to measure racial bias robustly under varying conditions.

Designing an ideal evaluation strategy is yet another crucial step in the face recognition processing pipeline. This step becomes particularly important in order to address racial bias within face recognition, as every decision made at this stage can have a significant impact on the overall performance and performance across different groups. In each decision, whether related to verification or identification tasks, there is a risk of misguiding the direction of research, particularly with regards to the development of face representation models, which can result in increased racial bias. Accordingly, we summarise the related literature addressing alternative evaluation methods within this stage and illustrate the corresponding stage and source of bias in Fig \ref{fig:overview}.

%% file: tex/5-Conclusions.tex
\vspace{-.4cm}
\section{Conclusions}
\label{sec:5}

We provide a comprehensive critical review of research on racial bias within face recognition. Firstly, we discuss the racial bias problem definition formalising the notions of the face recognition evaluation process and elucidate the prominent fairness criteria associated with face recognition. Subsequently, we highlight the racial grouping requirement of current fairness criteria and discuss standard race and race-related grouping terminology under three categories; race, skin tone and facial phenotypes and compare the most prominent grouping strategies across face recognition datasets. The high reliance of prior work on racial categories brings additional challenges as the race concept is defined and understood via the influence of pre-existing prejudices and discriminatory beliefs. Furthermore, skin tone remains only one trait of a comprehensive and multi-faceted race concept. Although a broader facial phenotype approach provides a more objective and granular evaluation strategy, ensuring that racial interpretations are not reduced to only facial phenotypes whilst also considering the broader context of historical and social factors, they remain important and under-explored research topics within the broader goal of achieving more accurate and fairer face recognition performance across increasingly more diverse populations.

Furthermore, we explore the contemporary automated facial recognition multiple-stage processing pipeline providing references to related work in the literature. In each stage, we cover the outline with a related baseline, standard procedures, a potential source of bias that can exacerbate racial bias and bias mitigation solutions. Firstly, the \textit{Image acquisition} stage consists of sources of bias (\textit{imagery bias, dataset bias}) that can affect the accuracy and fairness of face recognition systems. Such sources of bias within this initial stage will be transferred into the following stages and amplify racial bias in the final performance. Secondly, we consider the \textit{face localisation} stage in terms of racial bias, where there is little attention indicating the existence of racially disparate performance, but further investigation is explicitly needed targeting racial bias within face detection itself. Thirdly, we review the most fundamental works spanning the central stage of the face recognition pipeline, \textit{face representation}, under three sub-genres:- mutual information mitigation, loss function-based mitigation, and domain adaptation-based mitigation, providing an extensive supporting performance comparison across the RFW dataset. Finally, we investigate the final decision-making of the face recognition pipeline, \textit{face verification and identification} and reveal the impact of decision-making within this stage on overall and group-wise face recognition performance.

Overall we observe that racial bias is present at each and every technical stage of the face recognition pipeline such that the cumulative effect remains under-explored mainly in the literature. Furthermore, we observe continued bias within the evaluation strategies employed to measure the presence of this bias themselves that directly contradict the technological needs of a modern, diverse global society.

%% file: tex/references.tex
\small
\bibliographystyle{ACM-Reference-Format}
\bibliography{
references/base,
references/S2,
references/S3,
references/S4
}
